\RequirePackage{xcolor}
\documentclass{article}

\usepackage{microtype}
\usepackage{graphicx}
\usepackage{subfigure}
\usepackage{booktabs} %

\usepackage{hyperref}

\usepackage[commentsnumbered, longend]{algorithm2e}

\usepackage[accepted]{icml2021}

\usepackage{graphicx}
\usepackage{subfigure}  
\usepackage{todonotes}
\usepackage{dsfont}
\usepackage{tabulary}
\usepackage{tabularx}
\usepackage{makecell}
\usepackage{wrapfig}
\usepackage{microtype}      %
\usepackage{fancyvrb}

\usepackage{url}            %
\usepackage{amsfonts}       %
\usepackage{nicefrac}       %
\usepackage{xcolor}
\usepackage{amsmath,amsthm,amssymb}
\usepackage{mathtools}

\usepackage{etoolbox}
\usepackage{tikz}
\usetikzlibrary{arrows,automata,positioning}
\usetikzlibrary{shapes.geometric,shapes.symbols}

\usepackage{MnSymbol}%
\usepackage{wasysym}%
\usepackage{multirow} 
\usepackage{xspace}
\usepackage[capitalise]{cleveref}
\def\arrvline{\hfil\kern\arraycolsep\vline\kern-\arraycolsep\hfilneg}
\newcommand{\ie}[0]{\emph{i.e.},~}
\newcommand{\eg}[0]{\emph{e.g.},~}
\newcommand{\aka}[0]{a.k.a.~}

\newcommand{\defeq}{\ensuremath{\doteq}}

\usepackage{dsfont}
\usepackage{bm}

\newcommand{\mat}[1]{\ensuremath{\textbf{\MakeUppercase{{#1}}}}}

\newcommand{\set}[1]{\ensuremath{\mathbb{#1}}}
\newcommand{\gr}[1]{\ensuremath{{#1}}}

\newcommand{\ly}[1]{\ensuremath{^{(#1)}}}

\newcommand{\Reals}{\mathds{R}}
\newcommand{\Ns}{\ensuremath{\Omega}}%

\newcommand{\Vs}{[v]}
\newcommand{\Es}{E}
\newcommand{\Fs}{\Delta}

\newcommand{\Flags}{\Gamma}
\newcommand{\padding}{\Lambda}
\newcommand{\flag}{\gamma}
\newcommand{\Sg}{\ensuremath{\gr{Sym}}}
\newcommand{\Cg}{\gr{C}}
\newcommand{\Dg}{\gr{D}}
\newcommand{\Gg}{\gr{G}\xspace}
\renewcommand{\gg}{\gr{g}}
\newcommand{\Hg}{\gr{H}}
\newcommand{\Tg}{\gr{T}}
\newcommand{\Tgprod}{{\gr{T}}^{|\Fs|}}
\newcommand{\hg}{\gr{h}}
\newcommand{\Kg}{\gr{K}}

\newcommand{\kg}{\gr{k}}
\newcommand{\Ug}{\gr{U}}
\newcommand{\ug}{\gr{u}}
\newcommand{\sg}{\gr{s}}
\newcommand{\tg}{\gr{t}}

\newcommand{\Lm}{\mat{L}}
\newcommand{\Beta}{\mathrm{B}}

\newcommand{\gauge}{\ensuremath{^\text{gauge}}}
\newcommand{\hie}{\ensuremath{^\text{hierarchy}}}
\newcommand{\mainrho}{\ensuremath{^*}}

\newcommand{\bpi}{\boldsymbol{\pi}}
\newcommand{\brho}{\boldsymbol{\rho}}
\newcommand{\btau}{\boldsymbol{\tau}}
\newcommand{\bbeta}{\boldsymbol{\beta}}
\newcommand{\bsigma}{\boldsymbol{\sigma}}
\newcommand{\bkappa}{\boldsymbol{\kappa}}
\newcommand{\bupsilon}{\boldsymbol{\upsilon}}

\newtheorem{claim}{Claim}
\newtheorem{example}{Example}

\usepackage[framemethod=TikZ]{mdframed}

\mdfdefinestyle{MyFrame}{%
    linecolor=white,
    outerlinewidth=1pt,
    roundcorner=2pt,
    innertopmargin=\baselineskip,
    innerbottommargin=\baselineskip,
    innerrightmargin=5pt,
    innerleftmargin=5pt,
    backgroundcolor=black!5!white}

\mdfdefinestyle{MyFrame2}{%
    linecolor=black,
    outerlinewidth=1pt,
    roundcorner=3pt,
    innertopmargin=\baselineskip,
    innerbottommargin=\baselineskip,
    innerrightmargin=5pt,
    innerleftmargin=5pt,
    backgroundcolor=black!0!white}

\newbool{APX}
\booltrue{APX}

\icmltitlerunning{Equivariant Networks for Pixelized Spheres}

\begin{document}

\twocolumn[
\icmltitle{Equivariant Networks for Pixelized Spheres}

\begin{icmlauthorlist}
\icmlauthor{Mehran Shakerinava}{to,goo}
\icmlauthor{Siamak Ravanbakhsh}{to,goo}
\end{icmlauthorlist}

\icmlaffiliation{to}{School of Computer Science, McGill University, Montreal, Canada}
\icmlaffiliation{goo}{Mila - Quebec AI Institute}

\icmlcorrespondingauthor{Mehran Shakerinava}{mehran.shakerinava@mila.quebec}

\icmlkeywords{Machine Learning, ICML}

\vskip 0.3in
]

\printAffiliationsAndNotice{}  %

\begin{abstract}
Pixelizations of Platonic solids such as the cube and icosahedron have been widely used to represent spherical data, from climate records to Cosmic Microwave Background maps. Platonic solids have well-known global symmetries. Once we pixelize each face of the solid, each face also possesses its own local symmetries in the form of Euclidean isometries. 
 One way to combine these symmetries is through a hierarchy. However, this approach does not adequately model the interplay between the two levels of symmetry transformations.
We show how to model this interplay using ideas from group theory, identify the equivariant linear maps, and introduce equivariant padding that respects these symmetries. Deep networks that use these maps as their building blocks generalize gauge equivariant CNNs on pixelized spheres. 
These deep networks achieve state-of-the-art results on semantic segmentation for climate data and omnidirectional image processing.
Code is available at \url{https://git.io/JGiZA}.
\end{abstract}

\section{Introduction}
Representing signals on the sphere is an important problem across many domains; in geodesy and astronomy, discrete maps assign scalars or vectors to each point on the surface of the earth or points in the sky. To this end, various pixelizations or tilings of the sphere, often based on Platonic solids, have been used. Here, each face of the solid is refined using a
triangular, hexagonal, or square grid and further recursive refinements can bring the resulting polyhedron closer and closer to a sphere, enabling an accurate projection from the surface of a sphere; see \cref{fig:pixelizing}.

Our objective is to enable deep learning on this representation of spherical signals.
A useful learning bias when dealing with structured data is to design \emph{equivariant} models that preserve the symmetries of the structure at hand; the equivariance constraint ensures that symmetry transformations of the data result in the same symmetry transformations of the representation. 
To this end, we first need to identify the symmetries of pixelized spheres.

\begin{figure}
    \centering
    \includegraphics[width=1\linewidth]{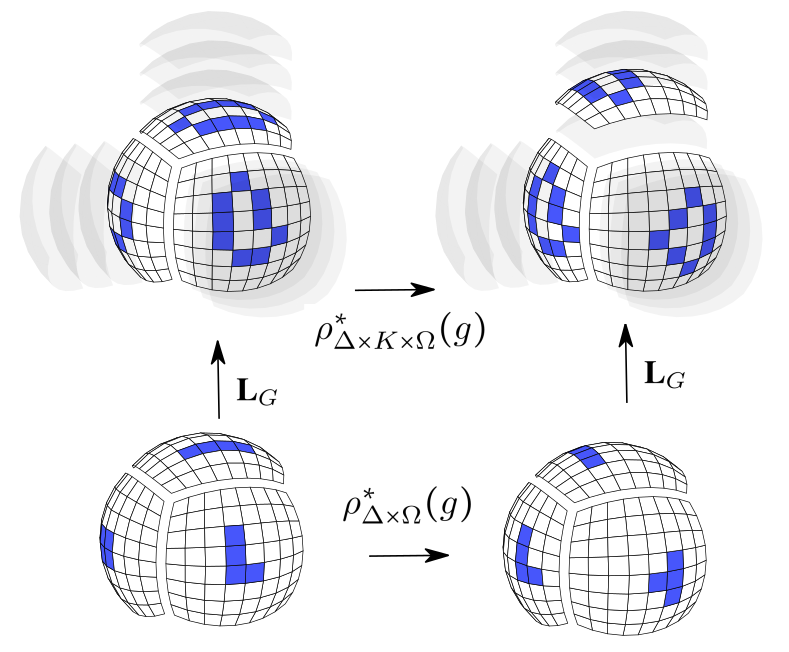}
    \vspace*{-2em}
    \caption{\footnotesize{We model the rotational symmetry of the sphere by combining the rotational symmetry of a Platonic solid and isometries of each of its face grids. The bottom row shows one such symmetry transformation for scalar features on the quad sphere. The top row shows the corresponding transformation for \emph{regular} features. Note that the $90^\circ$ rotation of the cube around the vertical axis also rolls the feature grids on top, in addition to rotating them. We identify the equivariant linear maps that make this diagram commute.}}
        \vspace*{-1em}
\end{figure}

While Platonic solids have well-known symmetries~\citep{coxeter1973regular}, their pixelization does not simply extend these symmetries. 
To appreciate this point it is useful to contrast the situation with the pixelization of a circle using a polygon: when using an $m$-gon, the cyclic group $\Cg_m$ approximates the rotational symmetry of the circle, $\gr{SO}(2)$. By further pixelizing and projecting each edge of the $m$-gon using $2$ pixels, we get a regular $2m$-gon, with a larger symmetry group $\Cg_m < \Cg_{2m} < \gr{SO}(2)$ -- therefore in this case further pixelization simply extends the symmetry.
However, this does not happen with the sphere and its symmetry group $\gr{SO}(3)$ -- that is, pixelized spheres are not homogeneous spaces for any finite subgroup of $\gr{SO}(3)$.

One solution to this problem proposed by \citet{cohen2019gauge} is to design deep models that are equivariant to ``local'' symmetries of a pixelized sphere. However, the symmetry of the solid is ignored in gauge equivariant CNNs. In fact, we show that under some assumptions, the gauge equivariant model can be derived by assuming a two-level hierarchy of symmetries~\citep{wang2020equivariant}, where the top-level symmetry is the complete exchangeability of faces (or local charts). 
A natural improvement is to use the symmetry of the solid to dictate valid permutations of faces instead of assuming complete exchangeability.

While the previous step is an improvement in modeling the symmetry of pixelized spheres, we observe that a hierarchy is inadequate because it allows for rotation/reflection of each face tiling independent of rotations/reflections of the solid. This choice of symmetry is too relaxed because the rotations/reflections of the solid completely dictate the rotation/reflection of each face-tiling. Using the idea of \emph{block systems} from permutation group theory, we are able to enforce inter-relations across different levels of the hierarchy, composed of the solid and face tilings.
After identifying this symmetry transformation, we identify the family of equivariant maps for different choices of Platonic solid. We also introduce an \emph{equivariant padding} procedure to further improve the feed-forward layer.

The equivariant linear maps are used as a building block in equivariant networks for pixelized spheres. Our empirical study using different pixelizations of the sphere demonstrates the effectiveness of our choice of approximation for spherical symmetry, where we report state-of-the-art on popular benchmarks for omnidirectional semantic segmentation and segmentation of extreme climate events.

\section{Pixelizing the Sphere}
To pixelize the sphere one could pixelize the faces of any polyhedron with transitive faces -- that is, any face is mapped to any other face using a symmetry transformation~\citep{popko2012divided}. Such a polyhedron is called an isohedron.
For example, the \emph{Quadrilateralized Spherical Cube} (quad sphere) pixelizes the sphere by defining a square grid on a cube. This pixelization was used in representing sky maps by the \emph{COsmic Background Explorer} (COBE) satellite. Alternatively, pixelization of the icosahedron using hexagonal grids for similar applications in cosmology is studied in \citet{tegmark1996icosahedron}. 
Today, a pixelization widely used to map the sky is \emph{Hierarchical Equal Area isoLatitude Pixelization} (HEALPix), which pixelizes the faces of a rhombic dodecahedron, an isohedron that is not a Platonic solid.

\begin{figure}[t]
  \centering
  \includegraphics[width=\linewidth]{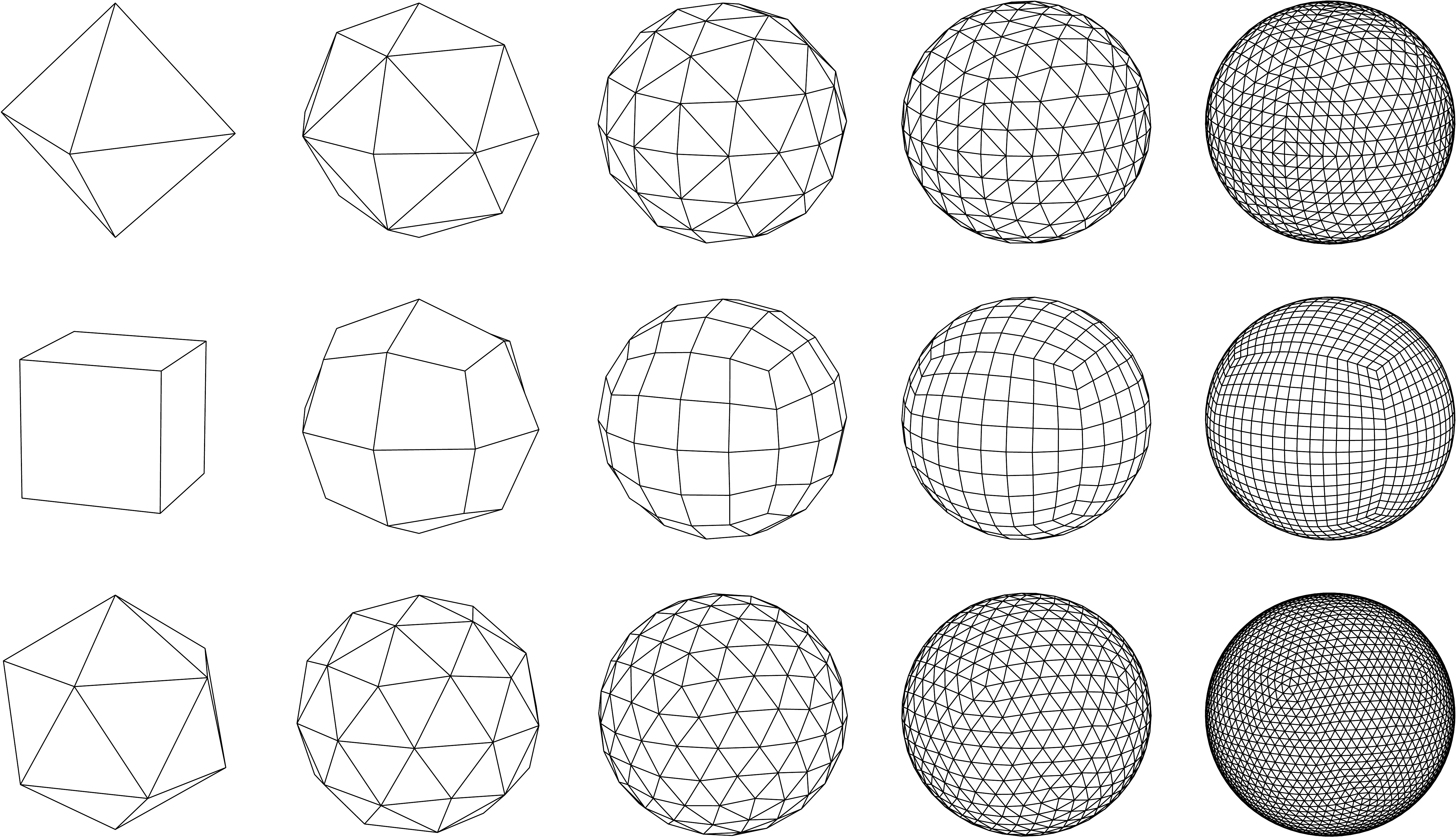}
    \vspace{-1.5em}
  \caption{\footnotesize{
   Iterative pixelization and projection for three Platonic solids:
   in each iteration (left-to-right), the pixels are recursively subdivided and projected onto the circumscribed sphere.
   }}
  \label{fig:pixelizing}
  \vspace{-1em}
\end{figure}

Platonic solids are more desirable as a model of the sphere because they are the only convex isohedra that are face-edge-vertex transitive -- that is, not only can we move any face to another face using symmetry transformations, but we can also do this for edges and vertices.
Similarly, there are only three regular \emph{tilings} of the plane with this property: triangular, hexagonal, and square grids.
Platonic solids give a regular tiling of the sphere, and this tiling is further refined by recursive subdivision and projection of each tile onto the sphere; see \cref{fig:pixelizing}.
A large family of \emph{geodesic polyhedra} use a triangular tiling to pixelize some Platonic solids, including the tetrahedron, octahedron, and icosahedron. In our treatment, we assume that rotation/reflection symmetries of each face
match the rotation/reflection symmetries of the tiling -- \eg square tiling is only used with a cube because both the square face of the cube and square grid have $90^\circ$ rotational symmetries. We exclude the dodecahedron because its triangular face tiling does not have translational symmetry.

\section{Preliminaries}
Let $[v] = \{1, \ldots, v\}$ denote the vertex set of a given Platonic solid. Each face $f \in [v]^m$ of the solid is an $m$-gon identified by its $m$ vertices, and $\Fs \subset [v]^m$ is the set of all faces. 
The action of the solid's symmetry group $\Hg$, \aka \emph{polyhedral group}, on faces $\Fs$ defines the permutation representation $\pi: \Hg \to \Sg(\Fs)$ that maps each group member to a permutation of faces. Here $\Sg(\Fs)$ is the group of all permutations of $\Fs$. We use $\pi(\Hg)$ to make this dependence explicit. Sometimes a subscript is used to identify
the \Hg-set -- for example, $\pi_\Fs(\Hg)$ and $\pi_{[v]}(\Hg)$ define $\Hg$ action on faces and vertices of the solid respectively. Since as a permutation matrix $\bpi(\hg): \Reals^{|\Fs|} \to \Reals^{|\Fs|}$ for $\hg \in \Hg$ is also a linear map, we use a bold symbol in this case to make the distinction.
For the same reason, we use $\Fs$ and $\Reals^{|\Fs|}$ interchangeably for the corresponding $\Hg$-set. 

\subsection{Symmetries of the Face Tiling}
Here, we focus on the symmetries of a single tiled face.
Each face has a regular tiling using a set $\Ns$ of \emph{tiles or pixels}. 
This regular tiling has its own symmetries, composed of 2D translations $\tau(\Tg) < \Sg(\Ns)$, and rotations/reflections $\kappa(\Kg) < \Sg(\Ns)$.\footnote{
One may argue that when the grid is projected to the sphere, the translational symmetry of the grid disappears since the grid is non-uniformly distorted. However, note that at the limit of having an infinitely high-resolution grid, this approximation (for small translations) becomes exact. Moreover, in a way, natural images also correspond to the projection of the 3D world onto a 2D grid, where we assume translational symmetry when using planar convolution.}
When we consider \emph{rotational} or \emph{chiral} symmetries $\Kg = \Cg_m$ is the \emph{cyclic group}, and when adding reflections, we have
 $\Kg = \Dg_m$, the \emph{dihedral group}. 
 
When combining translations and rotations/reflections one could simply perform translation followed by rotation/reflection. However, since a similar form of a combination of two transformations appears later in the paper (when we combine the rotations of the solid with translations on all faces), in the following paragraph, we take a more formal route to explain \emph{why} the combination of rotation/reflection and translation takes this simple form.

The rotation/reflection symmetries of the tiling define an \emph{automorphism} of translations
 $a: \Kg \to {Aut}(\Tg)$ -- \eg horizontal translation becomes vertical translation after a $90^\circ$ rotation. 
 This automorphism defines the \emph{semi-direct product} 
$\Ug = \Kg \rtimes_a \Tg$ as the abstract symmetry of the tiling.
The \emph{action} of members of this new group $(\kg, \tg) = \ug \in \Ug$,
on the tiles $\Ns$ is a permutation group $\upsilon(\Ug) < \Sg(\Ns)$
\begin{equation}\label{eq:face-action-scalar}
    \bupsilon_{\Ns}(\ug) = \underbrace{\left (\bkappa(\kg) \btau(\tg) {\bkappa(\kg^{-1})} \right )}_{\text{automorphism}\; a_\kg(\tg)} \bkappa{(\kg)} = \bkappa{(\kg)} \btau{(\tg)}.
\end{equation}
In the group action above, the automorphism of translations is through \emph{conjugation} by $\Kg$, where $\Kg$ itself also rotates/reflects the input. The end result becomes translation followed by rotation/reflection, as promised. 

The action above permutes individual pixels and therefore assumes \emph{scalar features} attached to each pixel.
An alternative is to define \emph{vector features} so that $\Ug$ action becomes \emph{regular}. For this we attach a vector of length $|\Kg|$ to each pixel, and use the regular $\Kg$ action on itself $\bkappa_\Kg: \Kg \to \Sg(\Kg)$ to define $\Ug$ action on the Cartesian product $\Kg \times \Ns$
\begin{equation}\label{eq:face-action-reg}
    \bupsilon_{\Kg \times \Ns}{(\kg,\tg)} = \bkappa_\Kg{(\kg)} \otimes \left( \bkappa_{\Ns}{(\kg)} \btau{(\tg)} \right),
\end{equation}
where $\otimes$ is the tensor product.
In words, $\Ug$ action on $\Kg \times \Ns$ translates and rotates/reflects the pixels $\Ns$ and at the same time transforms the vectors or \emph{fibers} $\Kg$.
 \begin{example}[Quad Sphere]\label{ex:1}
 Full symmetries of the cube is a subgroup of the orthogonal group $\Hg < \gr{O}(3)$. The corresponding rotations/reflections are represented by $3\times 3$ rotation/reflection matrices that have $\pm 1$ entries with only one non-zero per row and column. There are $2^3$ choices for the sign and $3! = 6$ choices for the location of these non-zeros, creating a group of size $6 \times 8 = 48$. Half of these matrices have a determinant of one and therefore correspond to rotational symmetries. For simplicity, in the follow-up examples, we consider only these symmetries. The resulting group is isomorphic to the symmetric group $\gr{S}_4$, where each rotation corresponds to some permutation of the four long diagonals of the cube.
 Now consider a $d \times d$ square tiling of each face of the cube, \ie $|\Ns(f)|=d^2$. 
In addition to translational symmetry $\Tg = \gr{C}_d \times \gr{C}_d$, the cyclic group $\Kg = \Cg_4$
represents the rotational symmetry of the grid. $\Ug$ action simply performs translation followed by  rotation in multiples of $90^\circ$.
\end{example}

\subsection{Equivariant Linear Maps for Each Face}
Given the permutation representations $\upsilon_{\Kg \times \Ns}(\Ug)$, a linear map 
$\Lm: \Reals^{|\Kg \times \Ns|} \to \Reals^{|\Kg \times \Ns|}$ is $\Ug$-equivariant if
    $\Lm \bupsilon(\ug) = \bupsilon(\ug) \Lm$ for all $\ug \in \Ug$, or in other words\footnote{Since $\bupsilon(\ug)$ is a permutation matrix, its inverse is equal to its transpose.} 
    $$\Lm = \bupsilon(\ug) \Lm  \bupsilon(\ug)^\top \quad \forall \ug \in \Ug.$$
    Using a tensor product property\footnote{$\operatorname{vec}(\mat{A}\mat{B}\mat{C}) = (\mat{C}^\top \otimes \mat{A}) \operatorname{vec}(\mat{B})$} we can rewrite this constraint as
    $$\operatorname{vec}(\Lm) =  \bupsilon^2(\ug) \operatorname{vec}(\Lm) \quad \forall \ug \in \Ug,$$
    where $\bupsilon^2(\ug) \defeq \bupsilon(\ug) \otimes \bupsilon(\ug)$ is a permutation action of $\ug \in \Ug$ on 
    $\set{A} = \Kg \times \Ns \times \Kg \times \Ns$, the elements of the ``weight matrix'' $\Lm$.
    The orthogonal bases for which this condition holds are $|\Kg \times \Ns| \times |\Kg \times \Ns|$ binary matrices $\Lm_{(1)}, \ldots, \Lm_{(\ell)}$ that are simply identified by the \emph{orbits} of $\upsilon^2(\Ug)$ action on $\set{A}$~\citep{wood1996representation,ravanbakhsh2017equivariance}. 
    The question of finding the linear bases is therefore the same as that of finding the orbits of permutation groups.
    We can use orbit finding algorithms from group theory with time complexity that is \emph{linear} in the number of input-outputs (\ie size of the matrix, or cardinality of $\set{A}$), and the size of the \emph{generating set} of the group, $\Gg^* \subseteq \Gg$ s.t. $\langle \Gg^* \rangle = \Gg$~\citep{hiss2007computational}. 
    Algorithm~1 in the Appendix gives the pseudo-code for finding the orbit  of a given element $a \in \set{A}$. 
    The fact that orthogonal bases are binary means that $\Ug$-equivariant linear maps are parameter-sharing matrices, where each basis identifies a set of tied parameters and its nonzero elements correspond to an orbit of $\Ug$ action on 
$\set{A}$. We have implemented this procedure for automated creation of parameter-sharing matrices and made the code available.\footnote{\fontsize{8pt}{10pt}\selectfont\url{https://github.com/mshakerinava/AutoEquiv}}

\begin{figure}
  \centering
  \begin{minipage}[c]{.47\linewidth}
    \includegraphics[width=\linewidth]{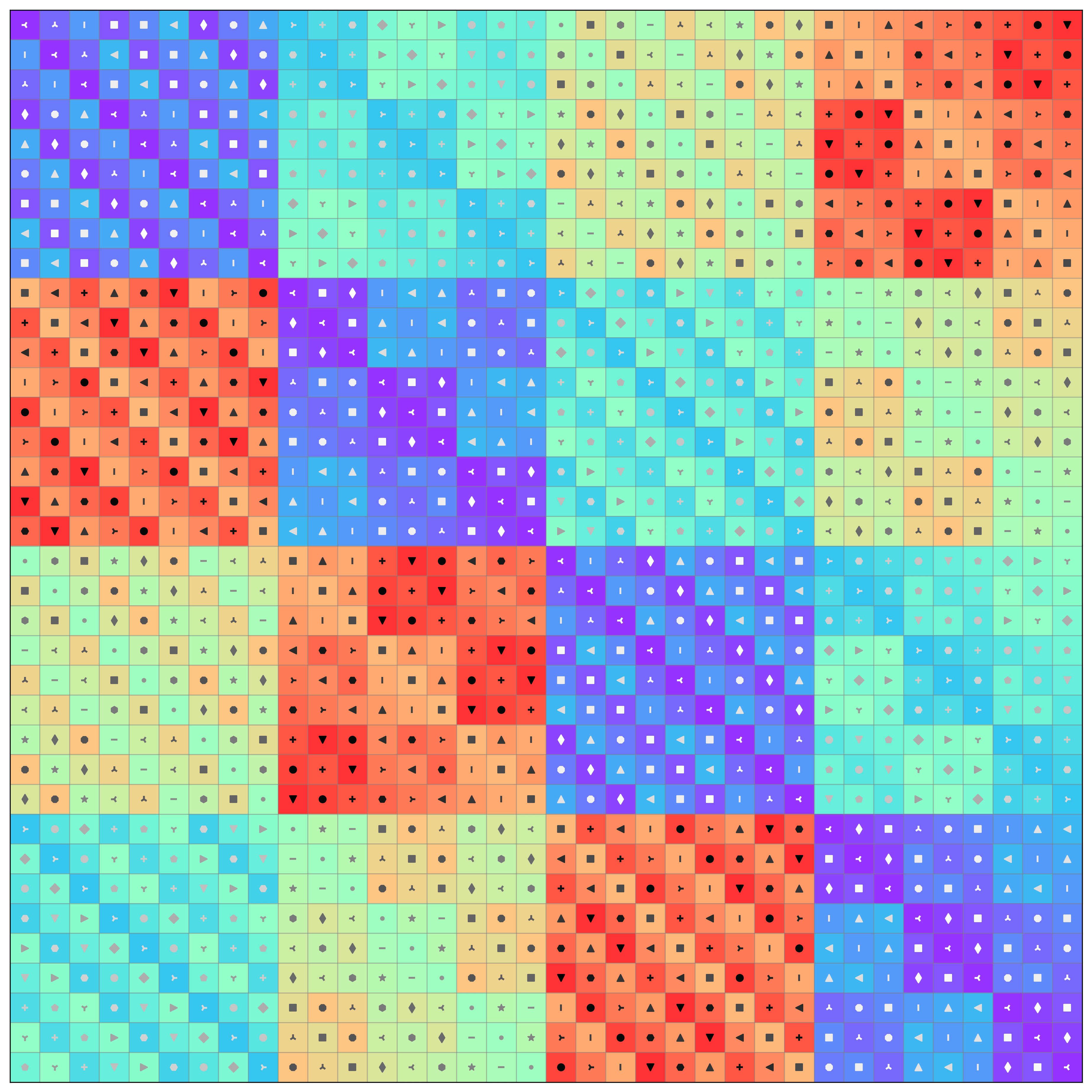}
  \end{minipage}\hfill
  \begin{minipage}[c]{0.49\linewidth}
  \vspace{-0.7 \baselineskip}
  \caption{\footnotesize{
  The parameter-sharing matrix for $\Lm_\Ug$. 
  This linear map is equivariant to circular translations and $90^\circ$ rotations of regular feature vectors on a $3 \times 3$ grid. That is, $\Kg=\Cg_4$ and $|\Ns| = 3\times3=9$. Note that this matrix is the parameter-sharing equivalent of having $\gr{C}_4$-steerable filters.
  }}
  \label{fig:Lu}
  \end{minipage}
  \vspace{-1.2em}
\end{figure}

To increase the expressivity of the deep network that deploys this kind of linear map, we may have multiple input and output \emph{channels}, and for each input-output channel pair, we use a new set of parameters.
An alternative characterization of equivariant linear maps
is through \emph{group convolution}, where the semi-direct product construction of $\Ug$
leads to $\Kg$-steerable filters~\citep{cohen2016group, cohen2016steerable}.
\cref{fig:Lu} gives an example of equivariant linear bases (parameter-sharing) for $3\times 3$ square tiling on each face of a quad sphere.

\begin{figure*}
  \centering
  \begin{minipage}[c]{.14\linewidth}
    \includegraphics[width=\linewidth]{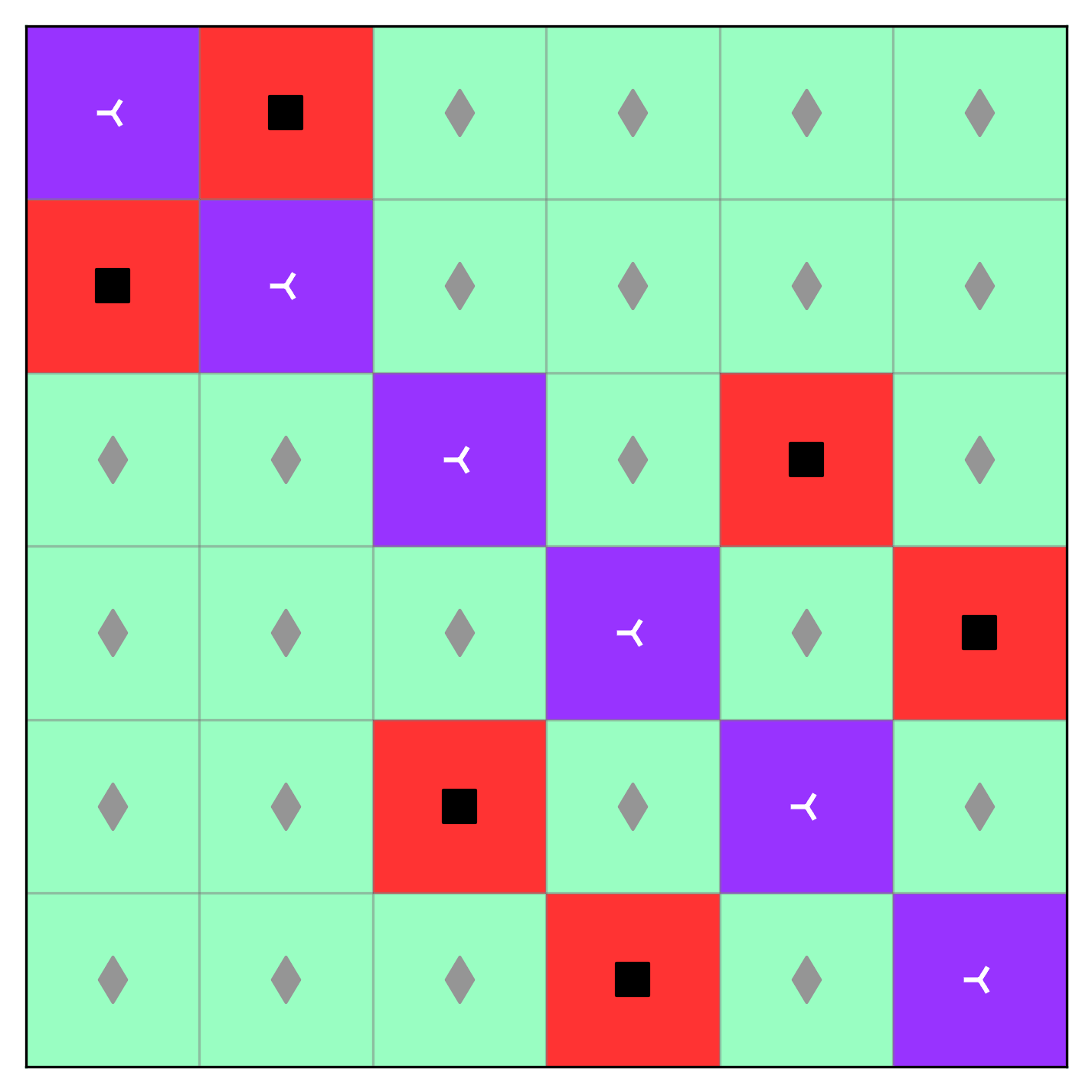}
  \end{minipage}\hfill
  \begin{minipage}[c]{0.83\linewidth}
  \vspace{-0.7 \baselineskip}
  \caption{\footnotesize{
 As we saw in \cref{ex:1} the rotational symmetries of the cube are given by $\Hg = \gr{S}_4$. 
The action of this group on the 6 faces is the permutation group $\pi_\Fs(\gr{S}_4)$. The parameter-sharing
constraint for $6\times 6$ matrices $\Lm_\Hg$ is shown in this figure. The corresponding $\Ug \wr \Hg = \Gg$-equivariant map $\Lm_\Gg\hie = \Lm_\Hg \otimes \left(1_{|\Kg \times \Ns|} 1_{|\Kg \times \Ns|}^\top\right) + \mat{I}_{|\Fs|} \otimes \Lm_\Ug$ assuming a $3\times 3$ face grid is constructed in two steps: 1) subdividing each row and column of $\Lm_\Hg$ into $3 \times 3 \times 4 = 28$ parts, to get a $216 \times 216$ matrix for $\Lm_\Hg \otimes \left(1_{|\Kg \times \Ns|} 1_{|\Kg \times \Ns|}^\top\right)$; 2) replacing the purple diagonal blocks with the $28\times28$ parameter-sharing matrix of \cref{fig:Lu}. This corresponds to the second term $\mat{I}_{|\Fs|} \otimes \Lm_\Ug$.}}
  \label{fig:Lh_face}
  \end{minipage}
  \vspace{-1em}
\end{figure*}

\section{Revisiting Gauge Equivariance}\label{sec:gauge}
\citet{cohen2019gauge} introduced gauge equivariant CNNs and used it to build Icosahedral CNN; an equivariant network for pixelization of the icosahedron. The idea is that a manifold as a geometric object may lack a global symmetry, and we can instead design models equivariant to the change of local symmetry, or \emph{gauge}. To establish the relationship between their model and ours, in their language, we assume each face to be a local chart\footnote{Note that \citet{cohen2019gauge} use several adjacent faces to create each chart and also associate the data with vertices rather than tiles. Moreover, our construction here ignores their $\Gg$-padding, and we discuss padding later. Our variation on their model makes some choices to help clarify what is missing in Icosahedral CNN.}. The interaction between local charts or faces is only through their overlap created by the padding of each face from adjacent faces. 
If we ignore the padding, their framework assumes an ``independent'' local transformation within each chart -- that is, their model is equivariant to independent translation and rotation/reflection within each face tiling. These independent transformations are represented by the product group
$\Ug_1 \times \ldots \times \Ug_{|\Fs|}$ acting on the set of all tiles $\Fs \times \Ns$, or $\Fs  \times \Kg \times \Ns$ in the case of regular features.
Furthermore, since the ``same'' model applies across charts, gauge equivariance assumes exchangeability of these transformations. The resulting overall symmetry group is, therefore, the \emph{wreath product} 
 \begin{equation}\label{eq:gauge-group}
     \Gg = \Ug \wr \Sg(\Fs)
 \end{equation}
in which a member of the symmetric group $\sg \in \gr{S}_{|\Fs|}$ permutes the transformations $\ug_1, \ldots, \ug_{|\Fs|} \in \Ug_1 \times \ldots \times \Ug_{|\Fs|}$.
Therefore, $\Gg$ action on $\Fs \times \Kg \times \Ns$ is given by the following 
\begin{align}\label{eq:gauge-group-action-reg}
     \brho_{\Fs \times \Ns}\gauge{(\gg)} = \left(\bsigma_\Fs{(\sg)} \otimes \mat{I}_{|\Kg \times \Ns|}\right) & \left (  \bigoplus_{f \in \Fs}  {\bupsilon_{\Kg \times \Ns}}{(\ug_f)} \right),
\end{align}
where $\gg = (\sg, \ug_1, \ldots, \ug_{|\Fs|})$. Here, the direct sum represents the independent transformation of each face by $\bupsilon(\ug_f)$ of \cref{eq:face-action-reg}, and the first term permutes the blocks in the direct sum using $\bsigma_\Fs(\sg)$. Action for scalar features simply replaces $\bupsilon_{\Kg \times \Ns}$ with $\bupsilon_{\Ns }$. %

\subsection{Gauge Equivariant Linear Map}
Previously we saw that $\Kg \rtimes \Tg = \Ug$-equivariant maps $\Lm_\Ug$ can be expressed using parameter-sharing linear layers. In~\citep{zaheer2017deep} it is shown that $\Sg(\Fs)$-equivariant maps take a simple form $\Lm_{\gr{S}(\Fs)} = w_1 \mat{I}_{|\Fs|} + w_2 (1_{|\Fs|} 1_{|\Fs|}^\top)$, where $\mat{I}_{c}$ is the $c\times c$ identity matrix, and 
$1_c = [1,\ldots,1]^\top$ is a column vector of length $c$. Given these components, as shown by~\citep{wang2020equivariant}, 
the equivariant map for the imprimitive action of their wreath product, as defined by \cref{eq:gauge-group-action-reg}
has the following form:
\begin{align} \label{eq:equivariant-gauge}
    \Lm_\Gg\gauge = \Lm_\gr{S} \otimes \left(1_{|\Kg \times \Ns|} 1_{|\Kg \times \Ns|}^\top\right) + \mat{I}_{|\Fs|} \otimes \Lm_\Ug.
\end{align}
In words, the resulting linear map applies the same $\Lm_\Ug$ to each face tiling, and
one additional operation pools over the entire set of pixels, multiplies the result by a scalar, 
and broadcasts back.
If we ignore the single global pool-broadcast operation, the result which simply applies an identical equivariant map to each chart coincides with the model of~\citep{cohen2019gauge}.

\section{Combining Local and Global Symmetries}
\subsection{Strict Hierarchy of Symmetries}\label{sec:hierarchy}
The symmetry group of \cref{eq:gauge-group} ignores the symmetries of the solid $\Hg$. However, adding $\Hg$ seems easy:
by simply replacing the representation $\sigma_\Fs(\gr{S})$ with $\pi_\Fs(\Hg)$ in \cref{eq:gauge-group-action-reg}, we get a smaller permutation group $\rho\hie(\Gg)$ acting on $\Fs \times \Kg \times \Ns$. 
Intuitively, this permutation group includes independent symmetry transformations of the tiling of each face while allowing the faces to be permuted according to the symmetries of the solid. The new permutation group is a subgroup of the old group: $\rho\hie(\Gg) < \rho\gauge(\Gg)$,
which means that the corresponding \Gg-equivariant map is less constrained or more expressive. The new \Gg-equivariant map $\Lm_\Gg\hie$ simply replaces $\Lm_{\gr{S}}$ in \cref{eq:equivariant-gauge} with $\Lm_\Hg$. Parameter-sharing layers equivariant to \Hg-action on faces $\bpi(\Hg)$ are easily constructed for different solids; see \cref{fig:Lh_face}.

While this approach is an improvement over the previous model, it is still inaccurate in the sense that it allows rotation/reflection of each face via $\kappa_\Ns(\Kg)$ independently of rotations/reflections of the
solid through $\pi_\Fs(\Hg)$. In principle, rotations/reflections of the solid completely determine the rotations/reflections of face tilings for all faces. Next, we find the symmetry transformation that respects this constraint and, by doing so, increase the expressivity of the resulting equivariant map.

\subsection{Interaction of Global and Local Symmetries}\label{sec:main-model}
Previously we observed that $\Hg$ action completely defines rotations and reflections of each face-tiling. Therefore our task is to define the pixelization symmetries $\Gg$ solely in terms of $\Hg$ and 
translations of individual tilings $\Tg$ (\ie we drop $\Kg$). 
Assuming independent translation within each face, we get the product group
\begin{equation}
   \Tgprod = \underbrace{\Tg\times \ldots \times \Tg}_{|\Fs| \; \mathrm{times}}.
\end{equation}
To combine this symmetry with the polyhedral symmetry $\Hg$ 
one should note that $\Hg$ itself acts on $\Tgprod$ -- \eg when we rotate the cube, translations are permuted and rotated. Geometrically, it is easy to see that $\Hg$ action on $\Tgprod$ is an automorphism -- there is a bijection between translations before and after rotating the solid. 
Let $b: \Hg \to {Aut}(\Tgprod)$ define an automorphism of $\Tgprod$ for each rotation/reflection  $\hg \in \Hg$ of the solid.  Then, the combined ``abstract'' symmetry is the \emph{semi-direct product} constructed using $b$ -- that is,
\begin{equation}\label{eq:main-global-sym}
    \Gg = \Hg \rtimes_b \Tgprod.
\end{equation}
Next, we define this group's permutation action on regular feature-fields $\Fs \times \Kg \times \Ns$ -- specialization to scalar fields is straightforward. To formalize this action, we need to introduce two ideas: 1) \emph{system of blocks} in permutation groups; 2) \emph{flags} and their properties in Platonic solids.

 \begin{figure*}[t!]
  \centering
  \begin{minipage}[c]{.16\textwidth}
    \includegraphics[width=\textwidth]{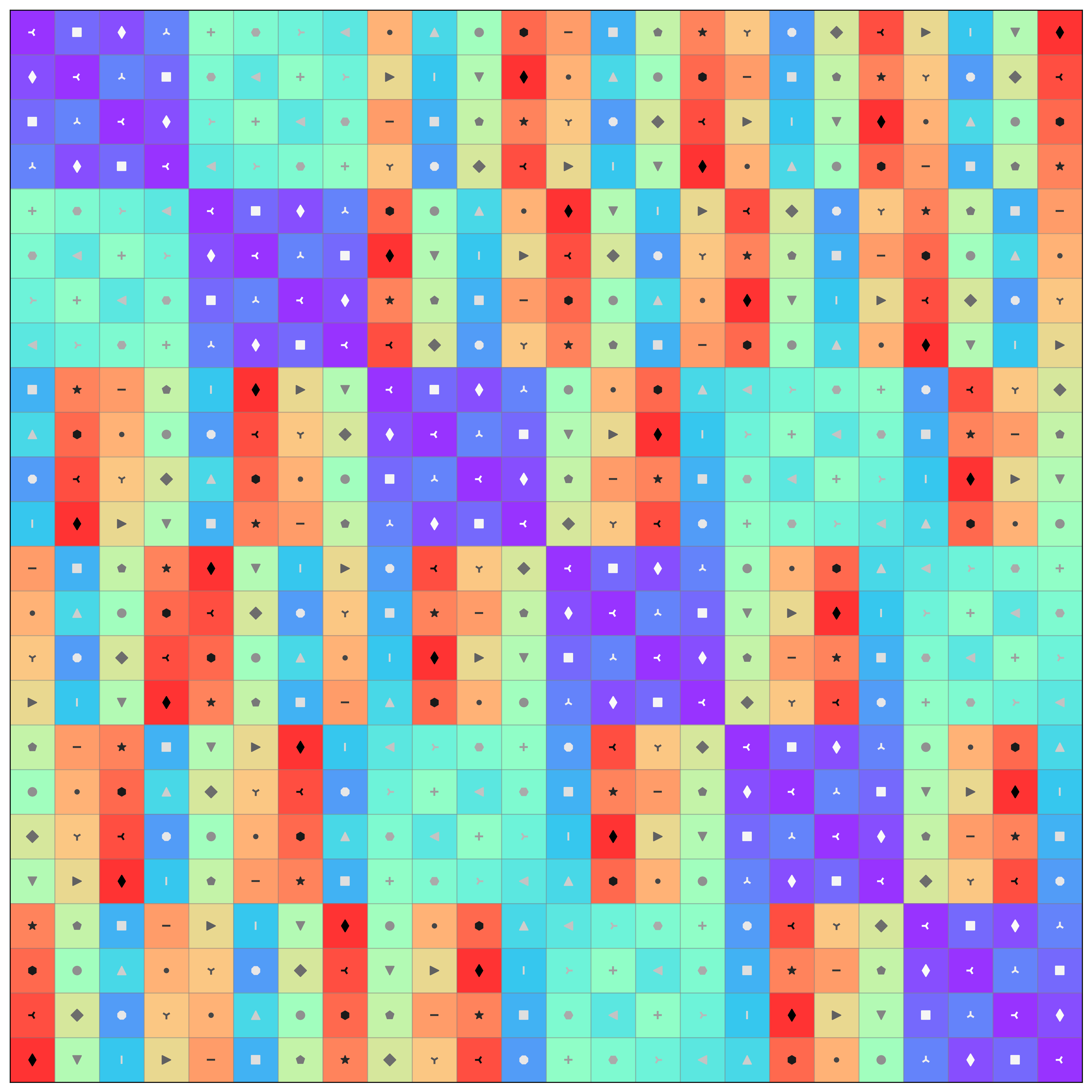}
  \end{minipage}\hfill
  \begin{minipage}[c]{0.81\textwidth}
  \vspace{-0.7 \baselineskip}
  \caption{\footnotesize{
  The parameter-sharing matrix for $\Lm_\Hg$, equivariant to $\pi_{\Fs \times \Kg}(\Hg)$ of \cref{eq:action-flags}; the rotations of the cube as they act on face-vertex pairs $\Flags_{\text{chiral}}$. Since the cube has $6\times4=24$ face-vertex pairs, this is a $24\times24$ matrix. To obtain the operations of \cref{eq:equivariant_main} for a quad sphere with $3\times 3$ tiling on each face, we need to repeat each row and column of this matrix $9$ ($=3\times 3$) times to get a $216\times216$ matrix corresponding to $\Lm_\Hg \otimes ({1}_{|\Ns|} {1}^\top_{|\Ns|})$. We then replace the $28\times 28$ purple blocks on the diagonal of the resulting matrix with identical copies of $\Lm_\Ug$ (\cref{fig:Lu}); this corresponds to the second term $\mat{I}_{|\Fs|} \otimes \Lm_\Ug$ in \cref{eq:equivariant_main}. The result is the parameter-sharing matrix of the equivariant map for a quad sphere, assuming regular features.
  }}
  \label{fig:Lflags}
  \end{minipage}
  \vspace{-1em}
\end{figure*}

\subsubsection{System of Blocks}
Consider $\rho(\Gg) < \Sg(\Theta)$, the permutation action of some $\Gg$ on a set $\Theta$. A block system is a partition of the set $\Theta$ into blocks $\Beta_1 \cupdot \ldots \cupdot \Beta_p$ such that the action of $\Gg$ preserves the block structure -- that is 
$(\gg \cdot \Beta) \cap \Beta$ is either $\emptyset$ or $\Beta$ itself, where the dot indicates the group action. 
 This means that for transitive sets we can identify the system of blocks using a single block $\Beta \subseteq \Theta$, and generate all the other blocks through $\Gg$ action.
 
Let $Stab_{\rho}(\theta) < \rho(\Gg)$ be the \emph{stabilizer subgroup} for $\theta \in \Theta$; this is the subset of permutations in $\rho(\Gg)$ that fix $\theta$. For the set 
$\Beta \subseteq \Theta$, let $Stab_{\rho}(\Beta) < \rho(\Gg)$ denote the \emph{set stabilizer} subgroup -- \ie
$\gg \cdot \Beta = \Beta$ for all $\gg \in Stab_{\brho}(\Beta)$. 
Given $\theta \in \Theta$, there is a bijection between subgroups of $\rho(\Gg)$ that contain $Stab_{\rho}(\theta)$
and systems of block $\Beta \ni \theta$~\citep{dixon1996permutation}. In other words, any block system $\Beta \ni \theta$ can be identified with its set-stabilizer $Stab_{\rho}(\Beta)$ which contains $Stab_{\rho}(\theta)$ as a subgroup.
Given a block system $\Beta \subseteq \Theta$, we can decompose the permutation matrices $\brho(\gg)$ as
\begin{equation}\label{eq:block-decomp}
    \brho(\gg) = \left (\brho_{\Theta / \Beta}(\gg) \otimes \mat{I}_{|\Beta|}\right) \left ( \bigoplus_{\Beta} \brho_{\Beta}(\gg) \right),
\end{equation}
where $\brho_{\Theta / \Beta}(\gg)$ permutes the blocks, and $\brho_{\Beta}(\gg) \in  Stab_\rho(\Beta)$, permutes the elements inside the block $\Beta$.
The reader may notice that the expression above resembles the wreath product action of \cref{eq:gauge-group-action-reg}. This is because the imprimitive action of the wreath product is a way of creating
block systems in which one group permutes the blocks and independent action of a second group permutes each inner block -- \ie these groups act independently at the two levels of the hierarchy.
However, to account for the interrelation between the global symmetry of the Platonic solid and the rotation/reflection symmetry
of each face tiling, we need to consider the system of blocks created by $\Hg$ action on \emph{flags}.

\subsubsection{Flags and Regular \Hg-Action} 
Adjacent face-edge-vertex triples of polyhedra are called \emph{flags}:
$
\Flags = \{ (f,e,v) \in \Vs \times \Es \times \Fs \mid  \{v\} \subset e \subset f \},
$
where $\Es \subset \Vs^2$ is the edgeset~\citep{cromwell1999polyhedra}.
An important property of Platonic solids is that their \emph{full symmetry} group $\Hg$ has a \emph{regular} action on flags -- \ie a unique permutation in the permutation group $\pi_\Flags(\Hg) < \Sg(\Flags)$ moves one flag to another. If we consider only the \emph{rotational} or \emph{chiral} symmetries, the group action is regular on adjacent face-vertex pairs $\Flags_{\mathrm{chiral}} = \{ (f,v) \in \Vs \times \Fs \mid  v \in f \}$. Moving forward, we work with flags, having in mind that our treatment
specializes to rotational symmetries by switching to $\Flags_{\mathrm{chiral}}$.

\subsubsection{\Gg-Action and Equivariant Map}
Now we have all the ingredients to define the $\Gg$ action, for $\Gg$ of \cref{eq:main-global-sym}, on 
the regular features of the pixelized sphere $\Fs \times \Kg \times \Ns$. 
The subset of flags associated with a face $\Flags(f) = \{(f,e,v) \in \Flags\} \subset \Flags$ 
 form a block system under $\Hg$ action -- that is rotations/reflections of the solid keep 
 the flags on the same face.  Moreover, the set-stabilizer subgroup ${Stab}_{{\pi_\Flags}}(\Flags(f))$ that fixes a face, is isomorphic to rotation/reflection symmetries of the face-tiling ${Stab}_{\pi_\Flags}(\Flags(f)) \cong \Kg$ and so
 it has a regular action on features $\Kg$.
Therefore, we can decompose $\Hg$ action on $\Flags$ as \cref{eq:block-decomp} 
 \begin{equation}\label{eq:action-flags}
     {\bpi}_\Flags(\hg) = \left (\bpi_\Fs(\hg) \otimes \mat{I}_{|\Kg|}\right) \left ( \bigoplus_{f \in \Fs} 
     \bpi_{\Flags(f)}(\hg) \right),
 \end{equation}
 where as before $\bpi_\Fs(\hg)$ permutes the faces, and $\bpi_{\Flags(f)}(\hg) \in Stab_{\pi_\Flags}(\Flags(f))$ is a permutation
 of the flags of face $f$. Since $Stab_{\pi_\Flags}(\Flags(f)) \cong \Kg$, each $\bpi_{\Flags(f)}(\hg)$ also uniquely
 identifies a rotation reflection of the tiling. Let $\bpi_{\Ns(f)}(\hg) < \Sg(\Ns)$ denote this action on the tiling.
 The combined action of the polyhedral group $\Hg$ on the pixelization is given by
 \begin{align*}
     \bbeta(\hg) = \left (\bpi_\Fs(\hg) \otimes \mat{I}_{|\Ns \times \Kg|}\right) 
     \left ( \bigoplus_{f \in \Fs} 
     \bpi_{\Flags(f)}(\hg)  \otimes \left (\bpi_{\Ns(f)}(\hg) \right) \right).
 \end{align*}
When defining the abstract symmetry of the solid we noted that the polyhedral group defines an automorphism of translations $b_h: \Tgprod \to \Tgprod$.
 $\beta(\Hg)$ concretely defines this automorphism through conjugation, resulting in the overall $\Gg$ action:
 \begin{align}
     & \brho\mainrho(\gg)  = \underbrace{\bbeta(\hg) \left (\bigoplus_f \btau(t_f) \otimes \mat{I}_{\Kg} \right ) \bbeta(\hg)^{-1}}_{\text{automorphism $b_\hg$}} \bbeta(\hg)  \label{eq:g-action-reg} \\
     & =  \left (\bpi_\Fs(\hg) \otimes \mat{I}_{|\Ns \times \Kg|}\right) 
      \left ( \bigoplus_{f \in \Fs} 
     \bpi_{\Flags(f)}(\hg)  \otimes \left (\bpi_{\Ns(f)}(\hg) \btau(\tg_f) \right) \right) \notag,
 \end{align}
where $\gg = (\hg, \tg_1, \ldots \tg_{|\Fs|})$. Here, $\beta(\Hg)$ transforms the translations, while also acting  on $\Fs \times \Kg \times \Ns$ (this is similar to \cref{eq:face-action-scalar}).
The end result is that 
 the combination $\bpi_{\Ns(f)}(\hg) \btau(\tg_f)$ performs translation followed by rotation/reflection
 on all tilings associated with a face, and $\bpi_{\Flags(f)}(\hg)$ permutes the tilings associated with flags of face $f$.
 
The general form of $\rho\mainrho$ is similar to $\rho\gauge$ of 
\cref{eq:gauge-group-action-reg}. The difference is that in that case we assumed
arbitrary shuffling of charts ($\sg \in \gr{S}$) as well as independent rotations/reflections of each face ($\kg_f \in \Kg$). Therefore we have $\gg = (\sg, (\tg_1, \kg_1), \ldots, (\tg_{|\Fs|}, \kg_{|\Fs|}))$. In our approach, $\sigma(\gr{S})$ is replaced by $\pi(\Hg)$, and all rotations/reflections $\kg_f$ for $f \in \Fs$ are dictated by $\bpi$ as well - therefore we have  $\gg = (\hg, \tg_1, \ldots \tg_{|\Fs|})$. 
As a permutation group we have 
$$\rho\mainrho < \rho\hie < \rho\gauge < \Sg(\Fs \times \Kg \times \Ns).$$

Next, we express the equivariant map for the pixelized sphere in terms of the equivariant map for the polyhedral group and the equivariant map for the face tiling.
\begin{mdframed}[style=MyFrame]
\begin{claim}\label{claim:1}
Let $\Lm_\Hg: \Reals^{|\Flags|} \to  \Reals^{|\Flags|}$ be equivariant to $\Hg$-action $\pi_\Flags(\Hg)$.
Similarly, assume $\Lm_{\Ug}: \Reals^{|\Kg \times \Ns|} \to \Reals^{|\Kg \times \Ns|}$ is equivariant to $\upsilon_{\Kg \times \Ns}(\Ug)$ as defined in \cref{eq:face-action-reg}. 
Then the linear map 
\begin{equation} \label{eq:equivariant_main}
    \Lm_\Gg = \Lm_\Hg \otimes ({1}_{|\Ns|} {1}^\top_{|\Ns|}) + \mat{I}_{|\Fs|} \otimes \Lm_\Ug
\end{equation}
 is equivariant to $\Gg$ action of \cref{eq:face-action-reg}.
 \end{claim}
 \end{mdframed}
 \vspace{-1em}
 The proof is in Appendix~A. Note that while the form of the equivariant map resembles the equivariant map for the hierarchy of symmetries (\eg \cref{eq:equivariant-gauge}), here we do not have a strict hierarchy; See \cref{fig:Lflags} for the example of the quad sphere.
  \begin{figure*}
  \centering
  \begin{minipage}[c]{.25\textwidth}
    \centering
    \includegraphics[width=0.48\linewidth]{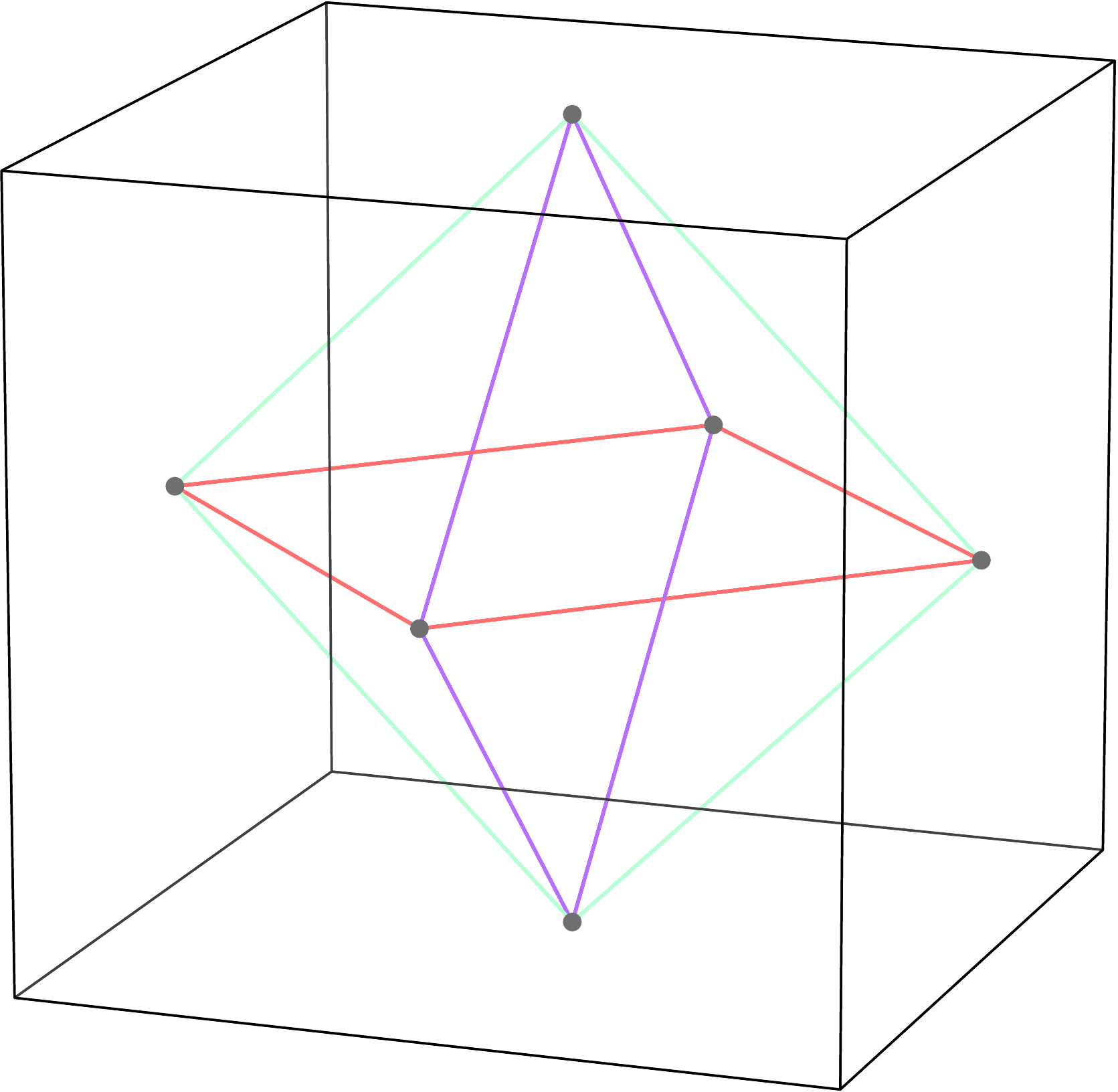}
    \hfill
    \includegraphics[width=0.48\linewidth]{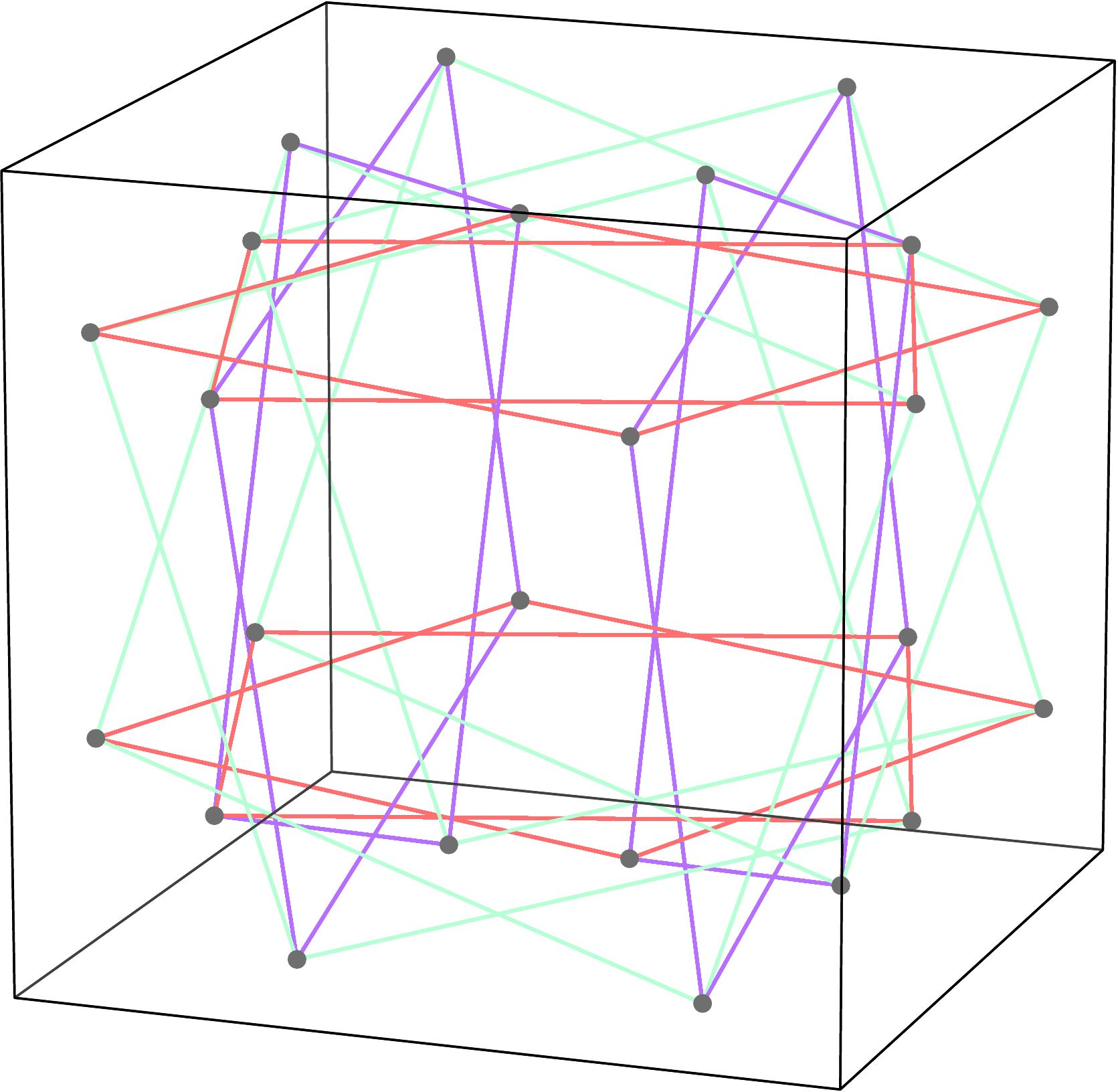}
  \end{minipage}\hfill
  \begin{minipage}[c]{0.7\textwidth}
  \vspace{-0.8 \baselineskip}
     \caption{\footnotesize{Padding graphs for our running example of quad sphere assuming rotational symmetry. Edge colors are added only to aid visualization.
    (\textbf{Left}) scalar features: the graph shows the neighborhood structure of faces. 
    (\textbf{Right}) regular features: each face-vertex pair (a face corner) identifies a $\flag \in \Flags_{\text{chiral}}$. In both cases, the graphs are invariant under the action of $\Hg$, which rotates the cube, where the group acts on the nodes of the padding graph.
    }}
    \label{fig:padding}
  \end{minipage}
  \vspace*{-1.5em}
\end{figure*}

\subsection{Orientation Awareness}\label{sec:orientation}
For some tasks, the spherical data may have a natural orientation. For example, in omnidirectional images, there is a natural up and down. In this case, equivariance to all rotations of the sphere over-constrains the model. A simple way to handle orientation in our equivariant map is to change $\Lm_\Hg$ to $\Lm_{\Hg'}$ where $\pi(\Hg') < \pi(\Hg)$ is the subgroup that corresponds to the desired symmetry. In the example of omnidirectional camera, when using a quad sphere, $\Hg'=\Cg_4$, and its action $\pi(\Hg')$ corresponds to rotations around the vertical axis.

\subsection{Equivariant Padding}\label{sec:padding}
Our goal is to define padding of face tilings for both scalar and regular features. 
For scalar features, it is visually clear which pixels are neighbors, and it is easy to produce such a padding operation. 
However, for regular features, where we have one tiling $\Ns(\flag)$ per flag $\flag \in \Flags$, padding is more challenging. Below we give a procedure.
Padding is a set of pairs $\padding \subset \Flags \times \Flags$ that identify neighboring flags. Since this neighborhood is symmetric, padding can also be interpreted as an undirected graph. Padding $\padding$ should satisfy the following two conditions: i) each pair belong to neighboring faces; ii) $\padding$ is equivariant to $\Hg$ action: 
\begin{align}
    adj(\padding) \; \bpi_{\Fs \times \Kg}(\hg) = \bpi_{\Fs \times \Kg}(\hg) \; adj(\padding) \quad \forall \hg \in \Hg,
\end{align}
where $adj(\padding) \in \{0,1\}^{|\Flags| \times |\Flags|}$ is the adjacency matrix of the padding graph. In other words,
$\pi(\Hg)$ defines the automorphisms of the padding graph; see \cref{fig:padding} for an example.
Our equivariant padding resembles \Gg-padding of~\citep{cohen2019gauge}; however, it is built using the high-level symmetry of the solid rather than relying on the choice of gauge in neighboring faces. 

\subsection{Efficient Implementation}\label{sec:efficient}
We build equivariant networks by stacking equivariant maps and $\mathrm{ReLU}$ nonlinearity: $\Lm_{\Gg}\ly{\ell} \circ \mathrm{ReLU} \ldots \circ \mathrm{ReLU}\circ \Lm_{\Gg}\ly{1}$. Invariant networks use additional global average pooling in the end.
For implementing $\Lm_\Gg$ of \cref{eq:equivariant_main}, we need efficient implementations of both terms in that equation. 
We implement the first term in \cref{eq:equivariant_main} using an efficient combination of \emph{Pool-Broadcast} operations:
$$\Lm_{\Hg} \otimes (1_{|\Ns|} 1^\top_{|\Ns|}) = \mathrm{Broadcast}_{\Ns} \circ \Lm_\Hg \circ \mathrm{Pool}_{\Ns}.$$
In words, we first pool over each tiling $\Ns(\flag) \; \forall \flag \in \Flags$, then apply the parameter-sharing layer $\Lm_\Hg$, and broadcast the result back to pixels $\Ns(\flag)$. To implement $\Lm_\Hg$, we use the parameter-sharing library mentioned earlier\footnote{\fontsize{8pt}{10pt}\selectfont\url{https://github.com/mshakerinava/AutoEquiv} is a library for efficiently producing parameter-sharing layers given the generators of any permutation group.}

\begin{figure}[ht]
  \centering
  \begin{minipage}[c]{.4\linewidth}
    \includegraphics[width=\linewidth]{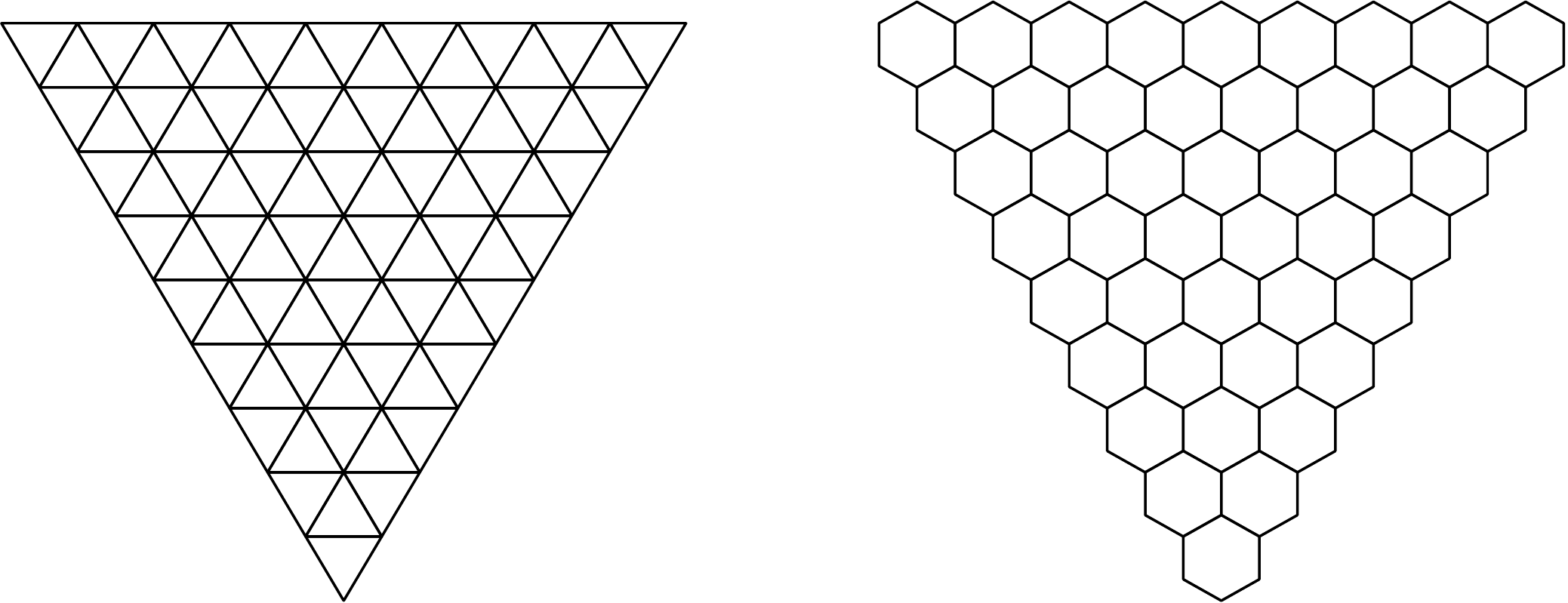}
  \end{minipage}\hfill
  \begin{minipage}[c]{0.55\linewidth}
  \vspace{-0.8 \baselineskip}
  \caption{\footnotesize{A triangular tiling is converted to a hexagonal
  tiling by replacing each down-facing triangle with a hexagon.}}
  \label{fig:hexa_face}
  \end{minipage}
  \vspace*{-1em}
\end{figure}

For triangular faces, we would like each pixel to be able to be translated to any other pixel. Therefore, we consider the input signal to lie only on downward-facing triangles. This is equivalent to considering a hexagonal grid of pixels; see \cref{fig:hexa_face}. We use implementations of group convolution for both square and hexagonal grids~\citep{cohen2016group, hoogeboom2018hexaconv}.

\section{Related Works}
Our contribution is related to a large body of work in equivariant and geometric deep learning. In design of equivariant networks one could either find equivariant linear bases~\citep{wood1996representation} or alternatively use group convolution~\citep{cohen2018general}; these approaches are equivalent~\citep{ravanbakhsh2020universal}. 
For permutation groups the first perspective leads to parameter-sharing
layers~\citep{ravanbakhsh2017equivariance} that are used in deep learning with sets~\citep{zaheer2017deep,qi2017pointnet}, tensors~\citep{hartford2018deep}, and graphs~\citep{kondor2018covariant,maron2018invariant}, where the focus has been on the symmetric group.
The group convolution approach which has been formalized in a series of works~\citep{cohen2016steerable,kondor2018generalization,cohen2018general,lang2020wigner} has been mostly applied to exploit subgroups of the Euclidean group. 
Among many papers that explore equivariance to Euclidean isometries are~\citep{marcos2017rotation,worrall2017harmonic,thomas2018tensor,bekkers2018roto,weiler20183d,weiler2019general}. Some of the papers that more specifically consider the subgroups of the orthogonal group are~\citep{cohen2018spherical,esteves2018learning,perraudin2019deepsphere,anderson2019cormorant,bogatskiy2020lorentz,dym2020universality}.
Other notable approaches include capsule networks~\citep{sabour2017dynamic,lenssen2018group}, 
equivariant attention mechanisms, and transformers~\citep{fuchs2020se,romero2020attentive,hutchinson2020lietransformer,romero2021group}.

The gauge equivariant framework of~\citep{cohen2019gauge} further extends the group convolution formalism,
and it has been applied to spherical data as well as 3D meshes~\citep{de2020gauge}. In addition to their model for the pixelized sphere, discussed in section \cref{sec:gauge}, the same framework is used to create a model for irregularly sampled points in \citet{kicanaoglu2020gauge}. Equivariant networks using global symmetries of geometric objects such as mesh and polyhedra that assume complete exchangeability of the nodes are studied in~\citep{albooyeh2019incidence}.

\begin{table*}[t!]
    \caption{\footnotesize{Results for different symmetry assumptions (no padding). 
    Pixel densities in figures match those of the experiments.}}\label{table:mnist-poly}
    \centering
    \scalebox{.83}{
    \begin{tabular}{l  c  c  c  c}
     & \textbf{Tetetrahedron} & \textbf{Cube} & \textbf{Octahedron} & \textbf{Icosahedron} \\
    \hfill $|\Fs|$ & 4 & 6& 8& 20 \\
    \hfill $|\Ns|$ & 861& 576& 325& 153 \\
    \hfill $\Hg$ & $A_4$ & $S_4$ & $S_4$ & $A_5$ \\
    & \includegraphics[width=.15\linewidth]{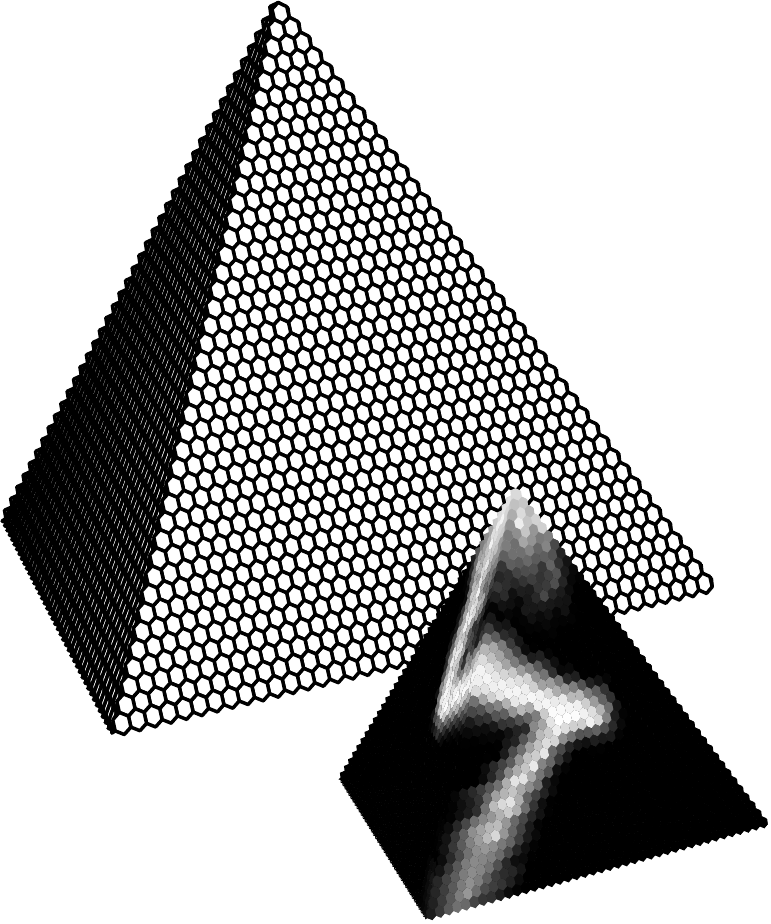} 
    & \includegraphics[width=.15\linewidth]{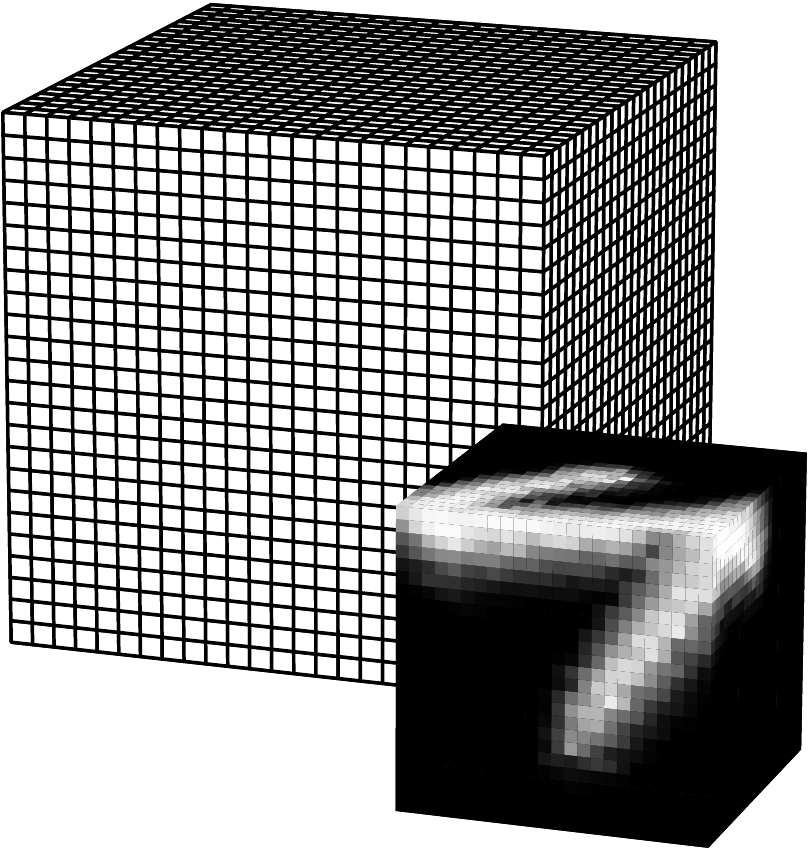} 
    & \includegraphics[width=.15\linewidth]{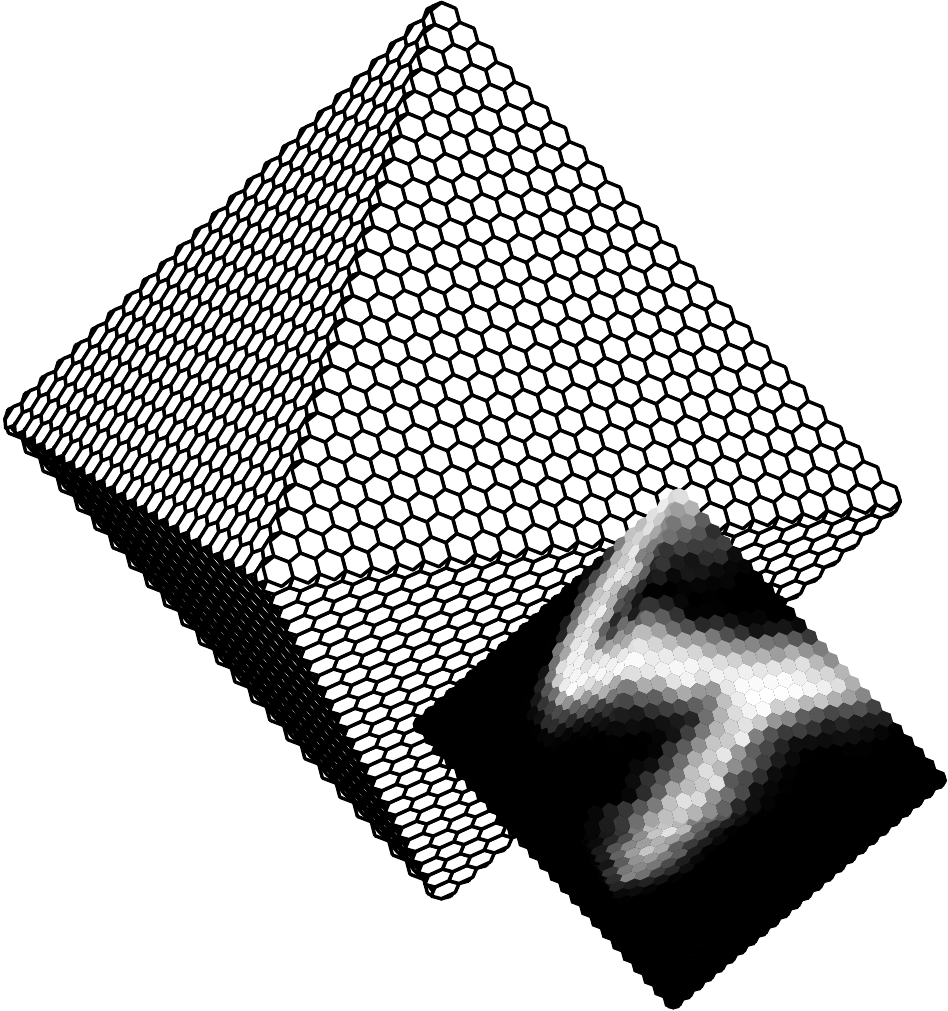} 
    & \includegraphics[width=.15\linewidth]{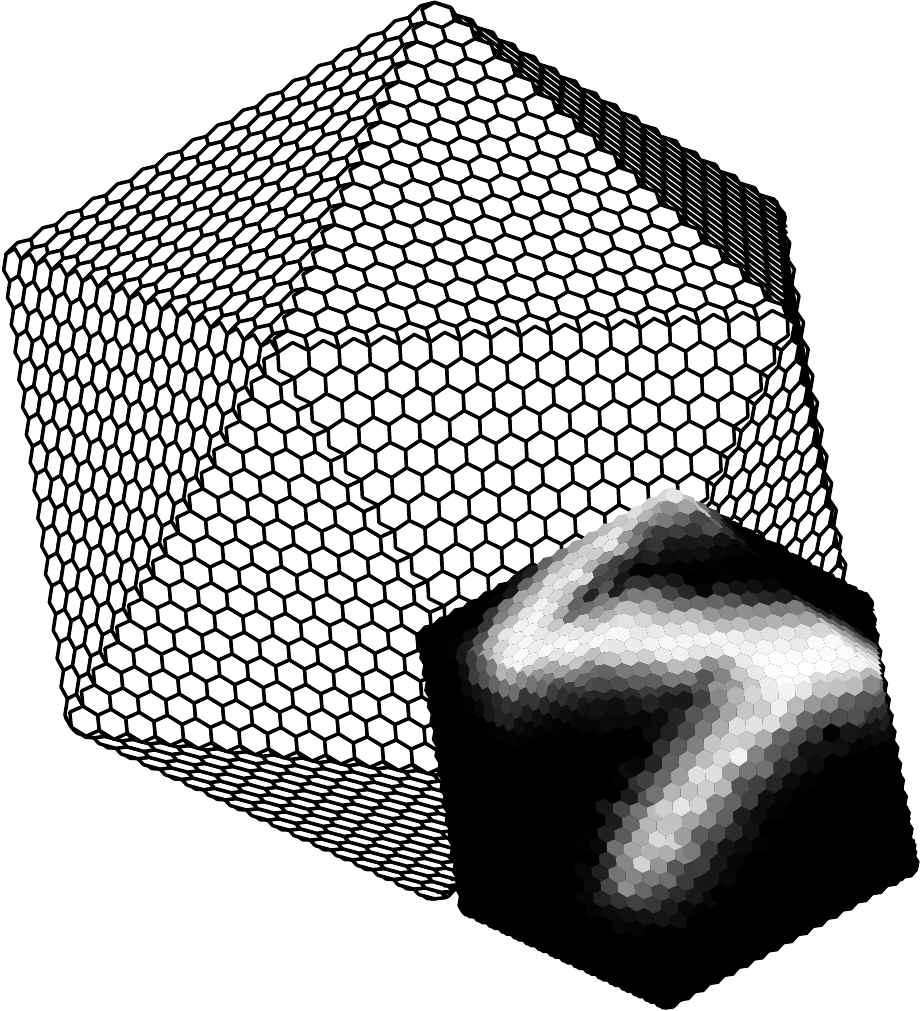}\\
    & \includegraphics[width=.15\linewidth]{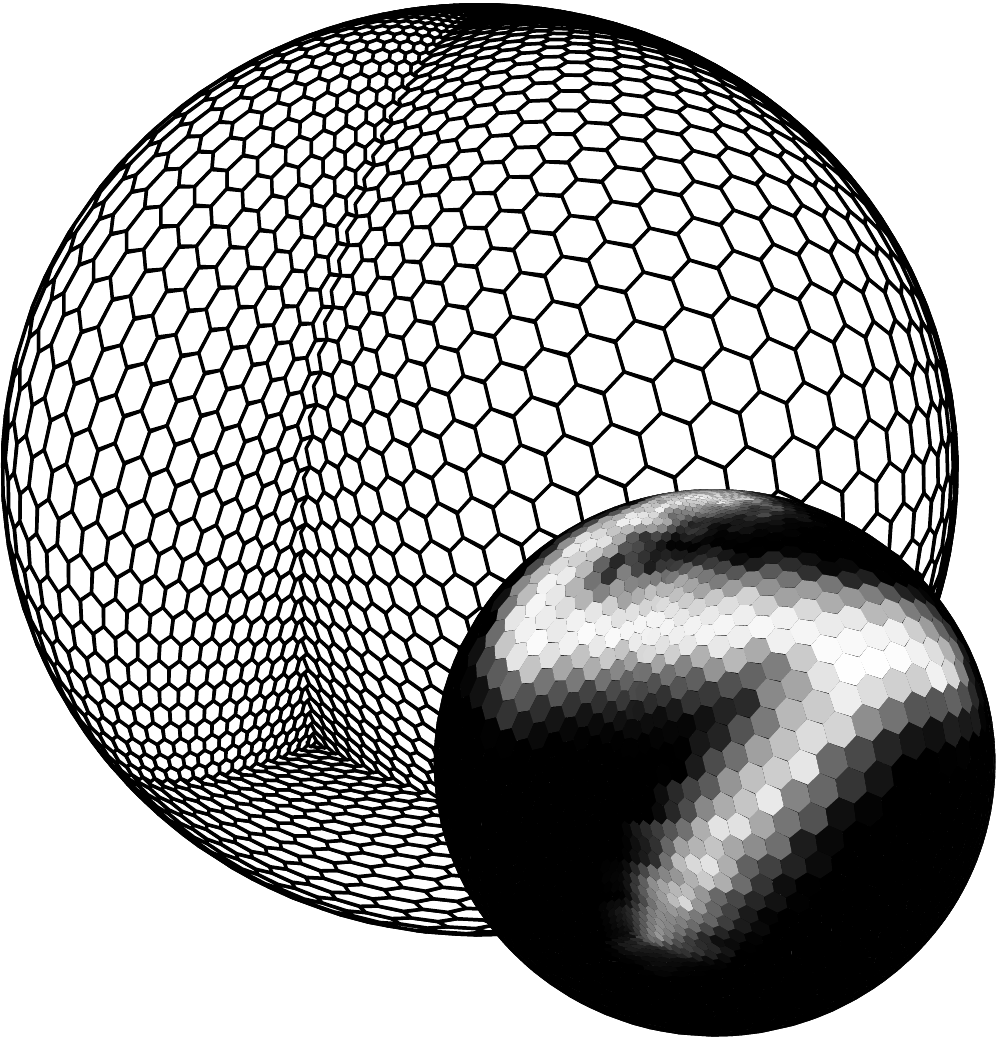} 
    & \includegraphics[width=.15\linewidth]{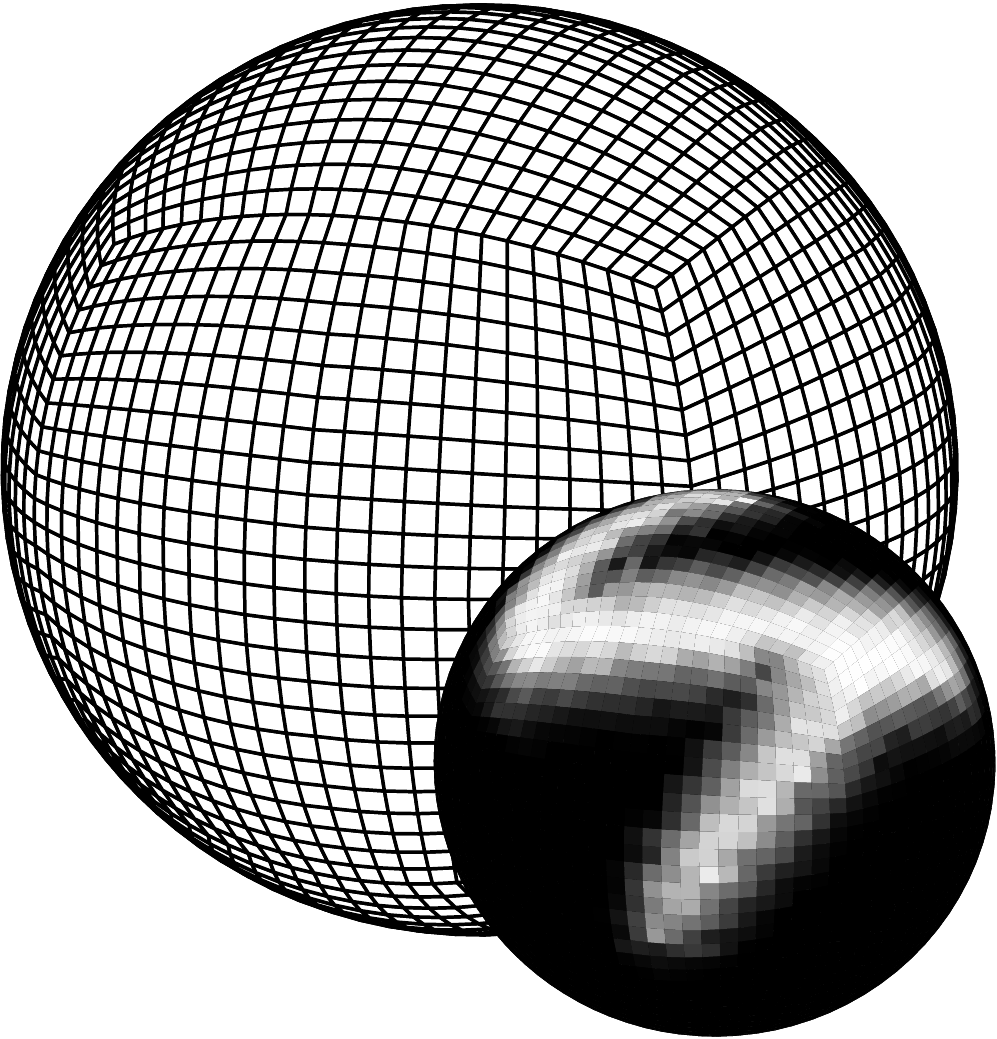} 
    & \includegraphics[width=.15\linewidth]{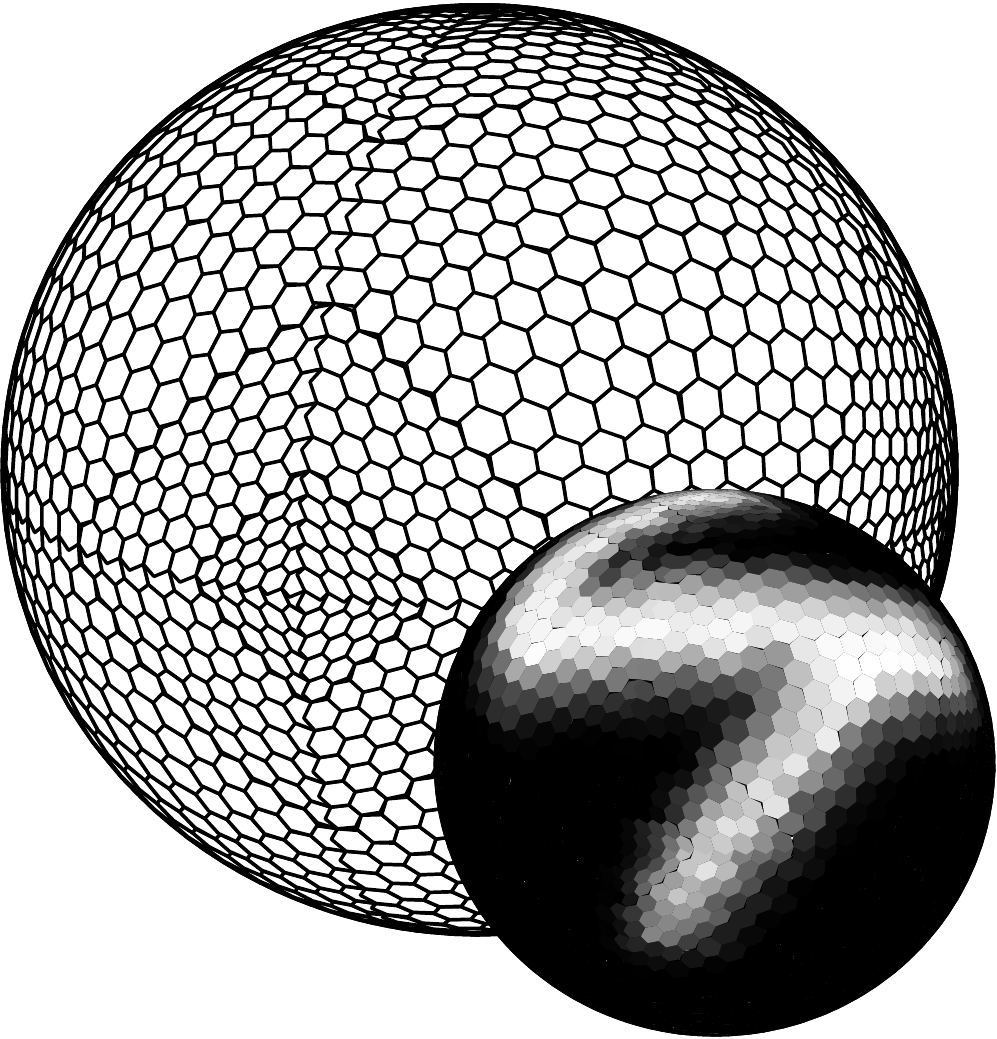} 
    & \includegraphics[width=.15\linewidth]{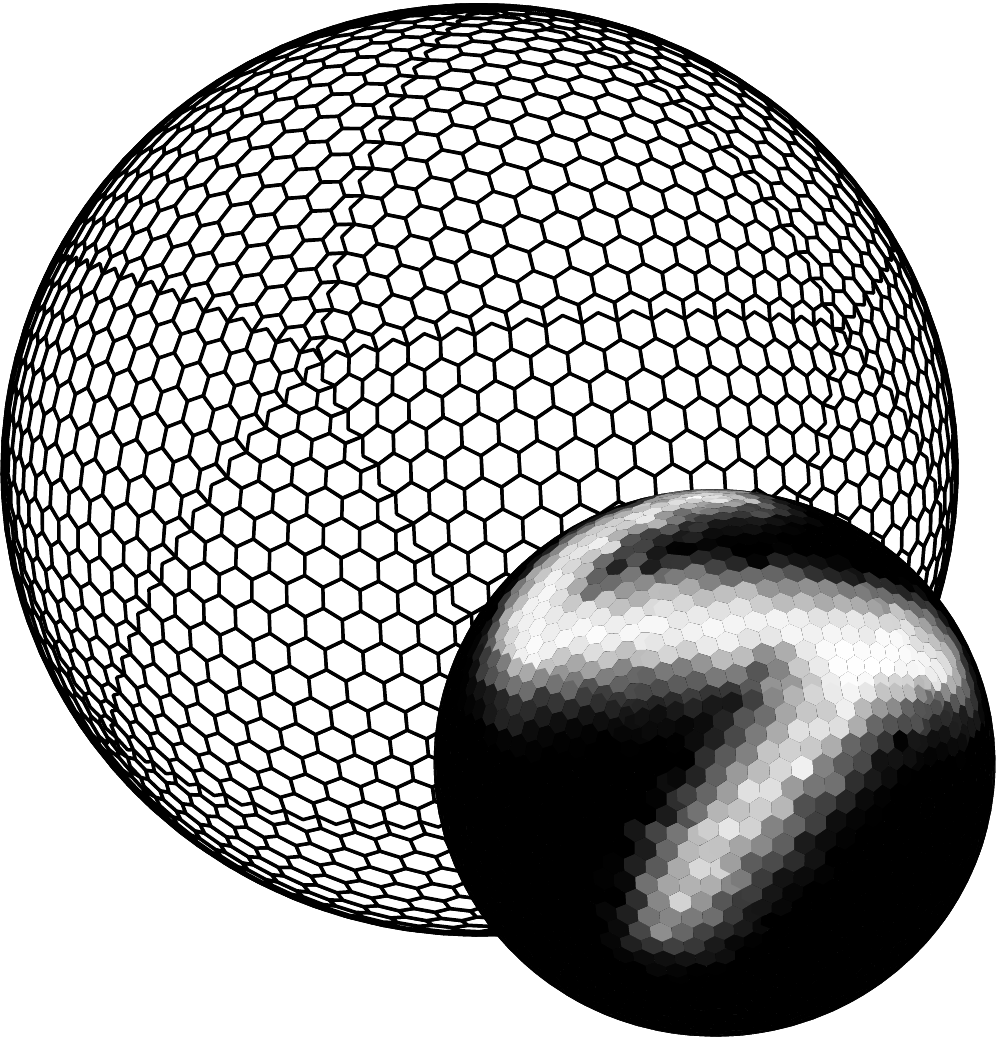}\\
    \multicolumn{5}{l}{\textbf{Using different number of channels for the global operations $\Lm_\Hg$ in the final model}}\\
    25\% & $98.79 \pm 0.03$ & $98.94 \pm 0.09$ & $98.96 \pm 0.08$ & $98.82 \pm 0.07$\\
    50\%  & $98.73 \pm 0.08$ & $98.99 \pm 0.05$ & $99.02 \pm 0.05$& $98.93 \pm 0.11$ \\
    75\% & $98.70 \pm 0.10$ & $98.88 \pm 0.04$ & $99.00 \pm 0.06$ & $98.76 \pm 0.10$ \\
    100\%  & $98.72 \pm 0.09$ & $98.83 \pm 0.05$& $98.95 \pm 0.07$ & $98.84 \pm 0.06$\\
    \multicolumn{5}{l}{\textbf{Comparison of different models}}\\
    Gauge ($\Ug \wr \gr{Sym}(\Delta)$) - \cref{sec:gauge} & $98.37 \pm 0.07$ & $98.70 \pm 0.06$ & $98.25 \pm 0.11$ & $97.45 \pm 0.12$\\
    Hierarchical ($\Ug \wr \Hg$) -  \cref{sec:hierarchy} & $98.62 \pm 0.10$ & $98.82 \pm 0.03$ & $98.62 \pm 0.03$ & $98.54 \pm 0.05$\\
    Our Final Model ($\Hg \rtimes \Tgprod$) - \cref{sec:main-model} & $98.73 \pm 0.08$ & $98.99 \pm 0.05$ & $99.02 \pm 0.05$& $98.93 \pm 0.11$ \\

    \end{tabular}
    }
\end{table*}

Below we quickly review other equivariant and non-equivariant models that are specialized for spherical data.
\citet{boomsma2017spherical} model the sphere as a cube and apply 2D convolution on each face with no parameter-sharing across faces.
\citet{su2017learning} and \citet{coors2018spherenet} design spherical CNNs for the task of omnidirectional vision. They use oriented convolution filters that transform according to the distortions produced by the projection method.
\citet{esteves2018learning} define a spherical convolution layer that operates in the spectral domain. Their model is $SO(3)$ equivariant, and their convolution filters are isotropic and non-localized.

\citet{cohen2018spherical} define an equivariant spherical correlation operation in $SO(3)$ which is further improved by
\citet{kondor2018clebsch} who introduce a Fourier-space nonlinearity and by \citet{cobb2021efficient} who make it more efficient. %
This enables them to implement the whole neural network in the Fourier domain.
\citet{jiang2019spherical} define convolution on unstructured grids via parameterized differential operators. They apply this convolution layer to the icosahedral spherical mesh (icosphere grid).
\citet{liu2018deep} introduce alt-az spherical convolution, which is equivariant to azimuth rotations, and implement it on the icosphere grid.

\citet{zhang2019orientation} model the sphere as an icosahedron. They unwrap the icosahedron and convolve it with a hexagonal kernel in two orientations. They then interpolate the result of the two convolutions to obtain a north-aligned convolution layer for the sphere.
\citet{defferrard2020deepsphere} build a graph on top of a discrete sampling of the sphere and then process this graph with an isotropic graph CNN. This is similar to using a gauge equivariant network with scalar feature maps.
\citet{esteves2020spin} define $SO(3)$ equivariant convolution between spin weighted spherical functions. Their convolutional filters are anisotropic and non-localized. Similar to \citet{esteves2018learning}, they make the filters more localized via spectral smoothness.

\begin{wraptable}{r}{.55\linewidth}
\vspace*{-5.5em}
\caption{\footnotesize{Spherical MNIST result.}}
\vspace{.2em}
\label{table:mnist}
\centering
\scalebox{.77}{
\begin{tabular}{lccr}
\toprule
Method & Acc. (\%) \\
\midrule
\citet{esteves2018learning} & 98.72 \\
\citet{cohen2018spherical}\footnotemark & 99.12 \\
\citet{cohen2019gauge} & 99.31 \\
\citet{esteves2020spin} & 99.37 \\
\citet{kicanaoglu2020gauge} & \textbf{99.43} \\
\textbf{Ours} (cube\ +\ padding) & 99.42 \\
\bottomrule
\end{tabular}
}
\vspace{-0.5em}
\end{wraptable}

\section{Experiments}
We use the efficient implementation of \cref{sec:efficient} in all experiments. Details of training and architectures appear in Appendix~D and Appendix~E. Below, we report our experimental results on spherical MNIST, Stanford 2D3DS, and HAPPI20 climate data.

\subsection{Spherical MNIST}
The spherical MNIST dataset~\citep{cohen2018spherical} consists of images from the MNIST dataset projected onto a sphere with random rotation. We consider the setting in which both training and test images are randomly rotated. As seen in \cref{table:mnist}, our quad sphere model is able to compete with state-of-the-art.\footnotetext{Taken from \citet{cohen2019gauge}}

Next, we study the effect of our global equivariant operation $\Lm_\Hg$ that depends on the symmetry of the solid, as well as the effect of the choice of solid. For these experiments, we do not use equivariant padding. We transform the dataset into our polyhedral sphere representation with the use of bilinear interpolation.
\cref{table:mnist-poly} visualizes different pixelizations and reports the choice of group $\Hg$ (for rotational symmetries) and the number of pixels per face $|\Ns|$.
Since the number of parameters in $\Lm_\Hg$ becomes large when we have multiple channels,
we allow for a fraction of channels to use $\Lm_\Hg$. The remaining channels only have local operations $\Lm_\Ug$. The first four rows of \cref{table:mnist-poly} show this fraction's effect on the performance.
Overall, we observe that using a fraction of channels for $\Lm_\Hg$ improves the model's performance. 

We then compare the performance of the three models discussed in \cref{sec:gauge,sec:hierarchy,sec:main-model}, where from top to bottom the size of the symmetry group decreases, and therefore we expect the model to become more expressive. The results are in agreement with this expectation.\footnote{Gauge model is equivalent to having $0\%$ global operations. For each of the remaining two models, we chose the best percentage of channels for global operations.}
Interestingly, in polyhedra with more faces, the effect of using more expressive global operations is generally more significant -- \eg for the icosahedron, the improvement is larger than that of the tetrahedron.

We choose to use the cube for the following experiments because of the simplicity and efficiency of its implementation and its good performance.

\begin{table}[t]
\caption{Results on Stanford 2D3DS.}
\label{table:2d3ds}
\begin{center}
\scalebox{.93}{
\begin{tabular}{l lcccr}
\toprule
& Method & \makecell{Mean  \\Acc. (\%)} & \makecell{Mean  \\IoU (\%)} \\
\midrule
\parbox[t]{2mm}{\multirow{4}{*}{\rotatebox[origin=c]{90}{\footnotesize{{non-oriented}}}}} & \citet{cohen2019gauge} & 55.9 & 39.4 \\
& \citet{kicanaoglu2020gauge} & 58.2 & 39.7 \\
& \citet{esteves2020spin} & 55.65 & 41.95 \\
& \textbf{Ours} (cube) & 58.74 & 40.99 \\
\midrule
\parbox[t]{2mm}{\multirow{3}{*}{\rotatebox[origin=c]{90}{\footnotesize{{oriented}}}}} & \citet{jiang2019spherical} & 54.7 & 38.3 \\
& \citet{zhang2019orientation} & 58.6 & 43.3 \\
& \textbf{Ours} (cube\ +\ orientation) & \textbf{62.5} & \textbf{45.0} \\
\bottomrule
\end{tabular}
}
\end{center}
\vspace*{-2em}
\end{table}

\subsection{Omnidirectional Camera Images}
The Stanford 2D3DS dataset~\citep{armeni2017joint} consists of 1413 omnidirectional RGBD images from 6 different areas. The task is to segment the images into 13 semantic categories.
We use the standard 3-fold cross-validation split and calculate average accuracy and IoU\footnote{Intersection over Union: $\frac{TP}{TP + FP + FN}$.} for each class over different splits. Then, we average the metrics obtained for each class to obtain an overall metric for this dataset.
We compose our equivariant map and equivariant padding in a U-Net architecture~\citep{ronneberger2015u}. Because of class imbalance, we use a weighted loss similar to previous works - \eg \citet{jiang2019spherical}. Our oriented model, described in \cref{sec:orientation}, achieves state-of-the-art performance on this dataset; See \cref{table:2d3ds}.

\subsection{Climate Data}
We apply our model to the task of segmenting extreme climate events~\citep{mudigonda2017segmenting}.
We use climate data produced by the Community Atmospheric Model v5 (CAM5) global climate simulator, specifically, the HAPPI20 run.\footnote{The data is accessible at {\fontsize{8pt}{10pt}\selectfont\url{https://p             ortal.nersc.gov/project/dasrepo/deepcam/segm_h5_v3_reformat}}.}
The training, validation, and test set size is 43917, 6275, and 12549, respectively.
The input consists of 16 feature maps.
We normalize each channel of the input to have zero mean and unit standard deviation.
The task is to segment atmospheric rivers (AR) and tropical cyclones (TC).
The rest of the pixels are labeled as background (BG).
See~\citet{mudigonda2017segmenting} for how ground truth labels are generated for this dataset.
The classes are heavily unbalanced with $0.1\%$ TC, $2.2\%$ AR, and $97.7\%$ BG.
To account for this unbalance, we use a \textit{weighted} cross-entropy loss.
We project the input features onto a quad sphere with $48 \times 48$ pixels/face. 
We use the non-oriented model for this task. 
In our experiments, we observed no significant gain from using the oriented model.
Experimental results can be seen in \cref{table:climate}.
Our method achieves the highest accuracy and average precision.
\footnotetext[11]{Result is taken from \citet{defferrard2020deepsphere}.}

\begin{table}[t]
\caption{Results on HAPPI20: mean accuracy (over TC, AR, BG) and mean average precision (over TC and AR).}
\label{table:climate}
\begin{center}
\vspace*{-1em}
\scalebox{.93}{
\begin{tabular}{lcc}
\toprule
Method & Mean Acc. (\%) & Mean AP (\%) \\
\midrule
\citet{jiang2019spherical}\footnotemark & 94.95 & 38.41 \\
\citet{zhang2019orientation} & 97.02 & 55.5 \\
\citet{cohen2019gauge} & 97.7 & 75.9 \\
\citet{defferrard2020deepsphere} & 97.8 $\pm$ 0.3 & 77.15 $\pm$ 1.94 \\
\citet{kicanaoglu2020gauge} & 97.1 & 80.6 \\
\textbf{Ours} (cube)  & \textbf{99.30} & \textbf{95.20} \\
\bottomrule
\end{tabular}
}
\end{center}
\vspace*{-1em}
\end{table}

\section*{Conclusion}
This paper introduces a family of equivariant maps for Platonic pixelizations of the sphere.
The construction of these maps combines the polyhedral symmetry with the rotation/reflection symmetry of their face-tiling. The latter is then, in turn, combined with the translational symmetry of pixels on each face to produce an overall permutation group.
Our use of \emph{system of blocks} to formalize this transformation merits further exploration as it provides a generalization of a hierarchy of symmetries.
Our derivation also demonstrates a close connection to gauge equivariant CNNs and suggests a generalization in settings where local charts possess a higher-level symmetry.
Our equivariant maps, which have efficient implementations, are combined with an equivariant padding procedure to build deep equivariant networks. 
These networks achieve state-of-the-art results on several benchmarks. Given the ubiquitous nature of spherical data, we hope that our contributions will lead to a broad impact. 

\section*{Acknowledgments}
We thank the anonymous reviewers for their valuable comments.
We would also like to thank the people at Lawrence Berkeley National Lab and Travis O'Brien for making the climate data available.
This project is supported by the CIFAR AI chairs program and NSERC Discovery. MS's research is in part supported by a
Kharusi Family International Science Fellowship.
Computational resources were provided by Mila and Compute Canada.

\newpage
\bibliography{refs}
\bibliographystyle{icml2021}

\clearpage
\appendix
\ifbool{APX}{\section{Proof}
\begin{proof} of \cref{claim:1} \\
 Our assumptions give us the following equivariance conditions:
 \begin{align}
     \Lm_\Ug \bupsilon_{\Kg \times \Ns}(\ug) & = \bupsilon_{\Kg \times \Ns}(\ug) \Lm_\Ug \quad  & \forall \ug \in \Ug \label{eq:p-ug}\\
     \Lm_\Hg \bpi_{\Fs \times \Kg}(\hg) & = \bpi_{\Fs \times \Kg}(\ug) \Lm_\Ug \quad  & \forall \hg \in \Hg
 \end{align}
  Our goal is to show that 
 \begin{equation}
      \brho\mainrho(\gg) \Lm_\Gg = \Lm_\Gg \brho\mainrho(\gg) \forall \gg \in \Gg,
 \end{equation}
for $\brho\mainrho(\gg)$ of \cref{eq:g-action-reg}. We do this in two parts.
 \paragraph{Part 1.} First we show $\mat{I}_{|\Fs|} \otimes \Lm_\Ug$ commutes with $ \brho\mainrho(\gg)$.
  From \cref{eq:p-ug}, using \cref{eq:face-action-reg}, and since $\Kg \cong Stab_{\pi_\Flags}(\Flags(f))$ we have $\forall \hg \in \Hg, f \in \Fs$
 \begin{align}
     & \Lm_\Ug  \left (\bpi_{\Flags(f)}(\hg)  \otimes \bpi_{\Ns(f)}(\hg) \btau(\tg_f)\right) = \notag \\ 
     & \quad \left (\bpi_{\Flags(f)}(\hg)  \otimes \bpi_{\Ns(f)}(\hg) \btau(\tg_f) \right) \Lm_\Ug \quad  \label{eq:p-ug-decomp}
 \end{align}
 Since $\bpi_{\Flags(f)}(\hg)  \otimes \bpi_{\Ns(f)}(\hg) \btau(\tg_f) = \bupsilon_{\Fs \times \Kg}(\ug)$ for some $\ug \in \Ug$, we use the short notation
 $$\bupsilon_{\Fs \times \Kg}(f,\hg) \defeq \bpi_{\Flags(f)}(\hg)  \otimes \bpi_{\Ns(f)}(\hg) \btau(\tg_f).$$
In this notation, we can rewrite $\brho\mainrho(\gg)$ as
 \begin{align}
     \brho\mainrho(\gg)  =  \left (\bpi_\Fs(\hg) \otimes \mat{I}_{|\Ns \times \Kg|}\right) 
     \left ( \bigoplus_{f \in \Fs} \bupsilon_{\Fs \times \Kg}(f,\hg)  \right) 
 \end{align}
Commutativity of $\mat{I}_{|\Fs|} \otimes \Lm_\Ug$ and $\brho\mainrho(\gg)$
follows from \cref{eq:p-ug-decomp} and the fact that the identity matrix commutes with 
any other matrix gives:
\begin{align*}
    &\overbrace{\left (\bpi_\Fs(\hg) \otimes \mat{I}_{|\Ns \times \Kg|}\right) 
     \left ( \bigoplus_{f \in \Fs} \bupsilon_{\Fs \times \Kg}(f,\hg)  \right)}^{\brho\mainrho(\gg)} 
     \left (\mat{I}_{|\Fs|} \otimes \Lm_\Ug \right)  \\
     = & \left (\bpi_\Fs(\hg) \otimes \mat{I}_{|\Ns \times \Kg|}\right) 
    \left (\mat{I}_{|\Fs|} \otimes \Lm_\Ug \right) 
  \left ( \bigoplus_{f \in \Fs} \bupsilon_{\Fs \times \Kg}(f,\hg)  \right)  \\
   = & \left (\mat{I}_{|\Fs|} \otimes \Lm_\Ug \right) \underbrace{\left (\bpi_\Fs(\hg) \otimes \mat{I}_{|\Ns \times \Kg|}\right) 
   \left ( \bigoplus_{f \in \Fs} \bupsilon_{\Fs \times \Kg}(f,\hg)  \right)}_{\brho\mainrho(\gg)}
\end{align*}
where in the second line, we used the fact that two block diagonal matrices with the same block sizes commute if corresponding blocks commute. To get the final equality we twice used the mixed product property of tensor product $(\mat{A} \otimes \mat{B})(\mat{C} \otimes \mat{D}) = (\mat{A}  \mat{C}) \otimes (\mat{B}  \mat{D})$:
\begin{align*}
    & \left (\mat{I}_{|\Fs|} \otimes \Lm_\Ug \right) \left (\bpi_\Fs(\hg) \otimes \mat{I}_{|\Ns \times \Kg|}\right)  \\
   = & \left (\mat{I}_{|\Fs|}  \bpi_\Fs(\hg)  \right) \otimes \left (  \Lm_\Ug \mat{I}_{|\Ns \times \Kg|} \right) \\
    = & \left ( \bpi_\Fs(\hg) \mat{I}_{|\Fs|}  \right) \otimes \left ( \mat{I}_{|\Ns \times \Kg|} \Lm_\Ug  \right) \\
    = &  \left (\bpi_\Fs(\hg) \otimes \mat{I}_{|\Ns \times \Kg|}\right) \left (\mat{I}_{|\Fs|} \otimes \Lm_\Ug \right)
\end{align*}

\paragraph{Part 2.} Next we show $\Lm_\Hg \otimes ({1}_{|\Ns|} {1}^\top_{|\Ns|})$ commutes with $\brho\mainrho(\gg)$.
For this, we need to decompose the second part of $\brho\mainrho(\gg)$ of \cref{eq:g-action-reg}. We use the following property of tensor product
$\mat{A} \otimes \mat{B} = (\mat{A} \otimes \mat{I}) (\mat{I} \otimes \mat{B})$, and the block-diagonal structure produced by direct sum to rewrite 
$\brho\mainrho(\gg)$ as 
\begin{align}
     \brho\mainrho(\gg) = &\overbrace{\left (\bpi_\Fs(\hg) \otimes \mat{I}_{|\Ns \times \Kg|}\right) \left ( \bigoplus_{f \in \Fs} 
     \bpi_{\Flags(f)}(\hg) \otimes \mat{I}_{|\Ns|} \right)}^{(i)} \\
     &\underbrace{\left ( \bigoplus_{f \in \Fs} \mat{I}_{|\Kg|} \otimes \left (\bpi_{\Ns(f)}(\hg) \btau(\tg_f) \right) \right)}_{(ii)}. \notag
\end{align}
Next, we show that each of the terms (i), and (ii) above commutes with $\Lm_\Hg \otimes ({1}_{|\Ns|} {1}^\top_{|\Ns|})$. First, rewrite (i) using \cref{eq:block-decomp} to get 
\begin{align}
  & \overbrace{\left ( \bpi_\Flags(\hg) \otimes \mat{I}_{|\Ns|} \right )}^{(i)}
    \left ( \Lm_\Hg \otimes ({1}_{|\Ns|} {1}^\top_{|\Ns|}) \right) \\
    = & \left ( \bpi_\Flags(\hg)   \Lm_\Hg ({1}_{|\Ns|} \right ) \otimes \left (  \mat{I}_{|\Ns|} ({1}_{|\Ns|} {1}^\top_{|\Ns|}) \right) \\
    = & \left ( \Lm_\Hg ({1}_{|\Ns|} \bpi_\Flags(\hg) \right ) \otimes \left (   ({1}_{|\Ns|} {1}^\top_{|\Ns|}) \mat{I}_{|\Ns|} \right) \\
    = & \left ( \Lm_\Hg \otimes ({1}_{|\Ns|} {1}^\top_{|\Ns|}) \right) \left ( \bpi_\Flags(\hg) \otimes \mat{I}_{|\Ns|} \right)
\end{align}
We then use the fact that (ii) is block-diagonal with $|\Fs \times \Kg|$ blocks, and $({1}_{|\Ns|} {1}^\top_{|\Ns|})$ commutes with any matrix to show commutativity:
\begin{align}
    & \left ( \bigoplus_{f \in \Fs} \mat{I}_{|\Kg|} \otimes \left (\bpi_{\Ns(f)}(\hg) \btau(\tg_f) \right) \right) \left ( \Lm_\Hg \otimes ({1}_{|\Ns|} {1}^\top_{|\Ns|}) \right) \\
    = &  \left ( \bigoplus_{f,\kg \in \Fs \times \Kg} \left (\bpi_{\Ns(f)}(\hg) \btau(\tg_f) \right) \right) \left ( \Lm_\Hg \otimes ({1}_{|\Ns|} {1}^\top_{|\Ns|}) \right ) \\
    = &  \left ( \Lm_\Hg \otimes ({1}_{|\Ns|} {1}^\top_{|\Ns|}) \right ) \left ( \bigoplus_{f,\kg \in \Fs \times \Kg} \left (\bpi_{\Ns(f)}(\hg) \btau(\tg_f) \right) \right )\\
    &  \left ( \Lm_\Hg \otimes ({1}_{|\Ns|} {1}^\top_{|\Ns|}) \right) \left ( \bigoplus_{f \in \Fs} \mat{I}_{|\Kg|} \otimes \left (\bpi_{\Ns(f)}(\hg) \btau(\tg_f) \right) \right).
\end{align}

Parts 1. and 2. together complete the proof. 
\end{proof}

\newpage
\section{Equivariant Padding for the Cube}\label{sec:padding_appendix}

\cref{fig:padding_face_simple} shows how a face of the cube is padded when we have scalar features. \cref{fig:padding_face} shows equivariant padding of a face in the more challenging case of regular features. Here we have one tiling $\Ns(\flag)$ for each $\flag \in \Flags_{\text{chiral}}$ -- \ie one
regular grid per face-vertex pair. We have visualized these grids by putting them close to a corner.
We pad the corner pixels with zero.
\begin{figure}[h]
    \centering
    \includegraphics[width=0.38\linewidth]{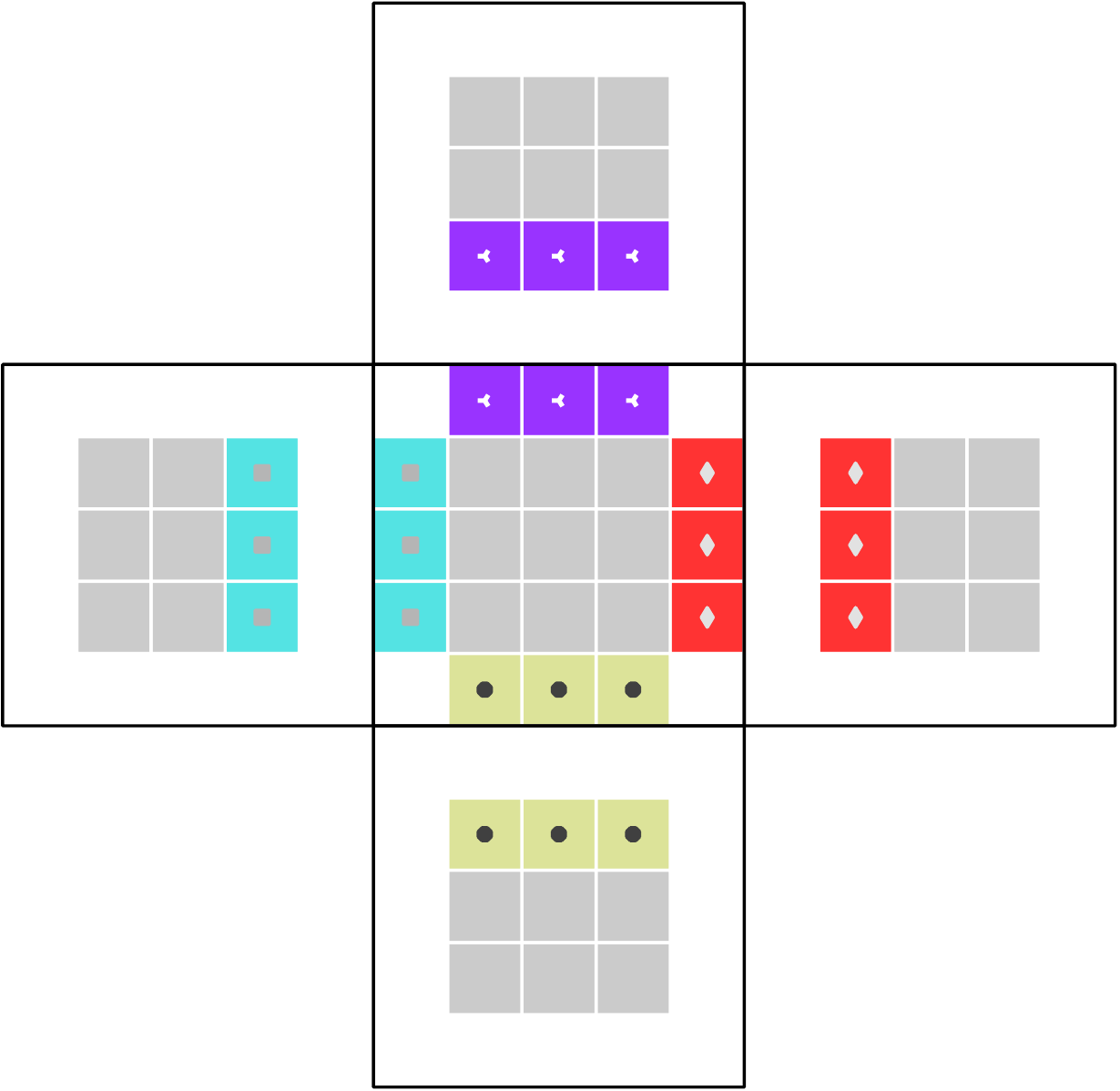}
    \caption{
        Equivariant padding applied to a single face of the cube in the setting where there is one grid per face.
    }
    \label{fig:padding_face_simple}
\end{figure}

\begin{figure}[h]
    \centering
    \includegraphics[width=0.76\linewidth]{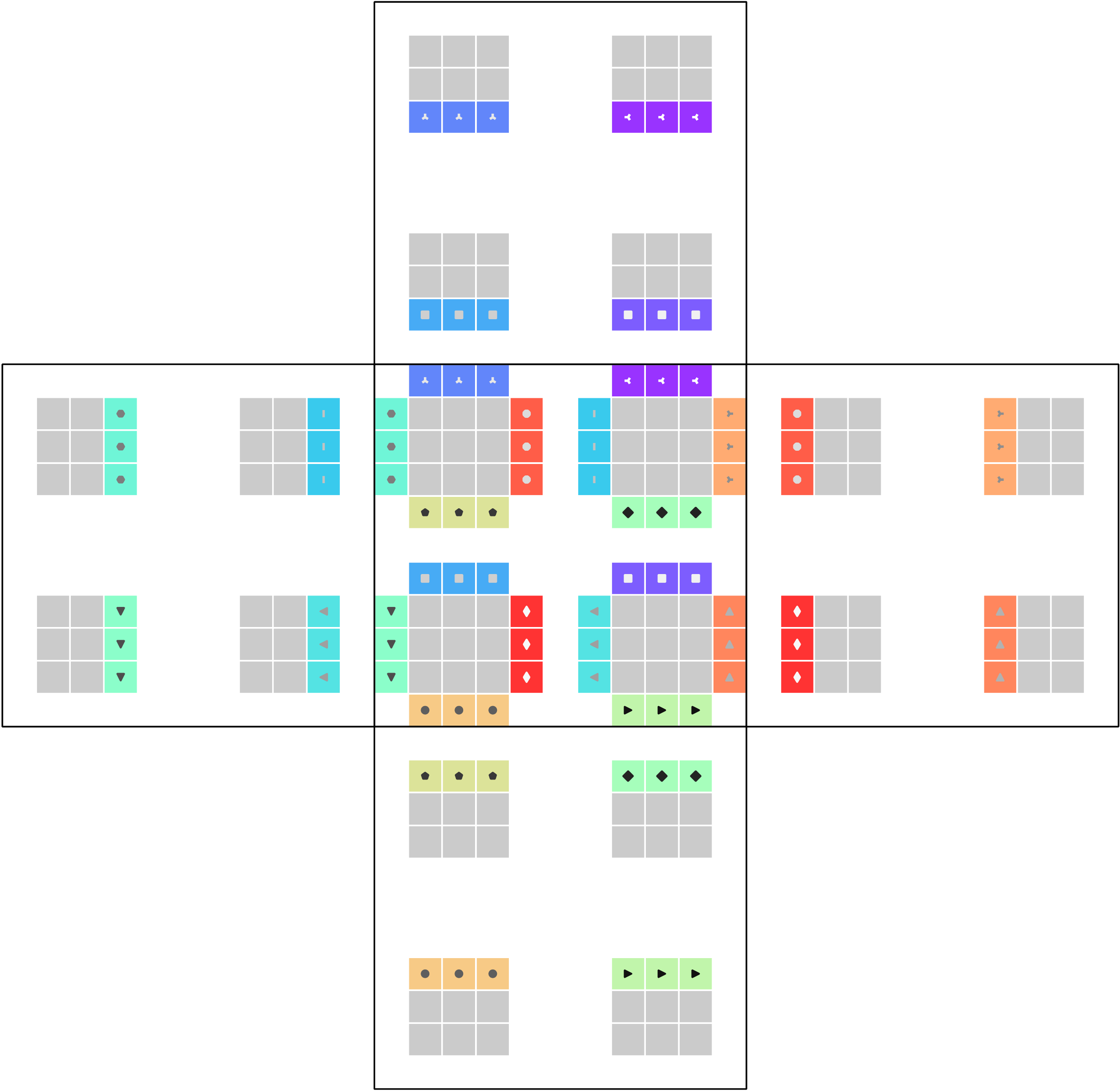}
    \caption{
        An example of an equivariant padding applied to a single face of the cube. The top left feature grid is matched to the top left feature grids of neighboring faces, the top right feature grid is matched to the top right feature grids of neighboring faces, and so on. The complete padding operation consists of applying this operation to each face of the cube. One can verify equivariance by noting that rotation followed by padding is equivalent to padding followed by rotation.
    }
    \label{fig:padding_face}
\end{figure}

\section{Hexagonal Pooling}\label{sec:hexa_pool}

It is common for CNNs to include pooling layers that reduce the spatial dimension of internal representations.
\cref{fig:hexa_pool} shows how the pooling layer operates on a hexagonal grid. Pooling changes the width from $w$ to $(w + 1)/2$. Since there are three pooling layers in our neural network architectures (see \cref{sec:arch}), the initial width of the hexagonal grid has to be of the form $8k + 1$.

\begin{figure}[h]
    \centering
    \includegraphics[width=0.52\linewidth]{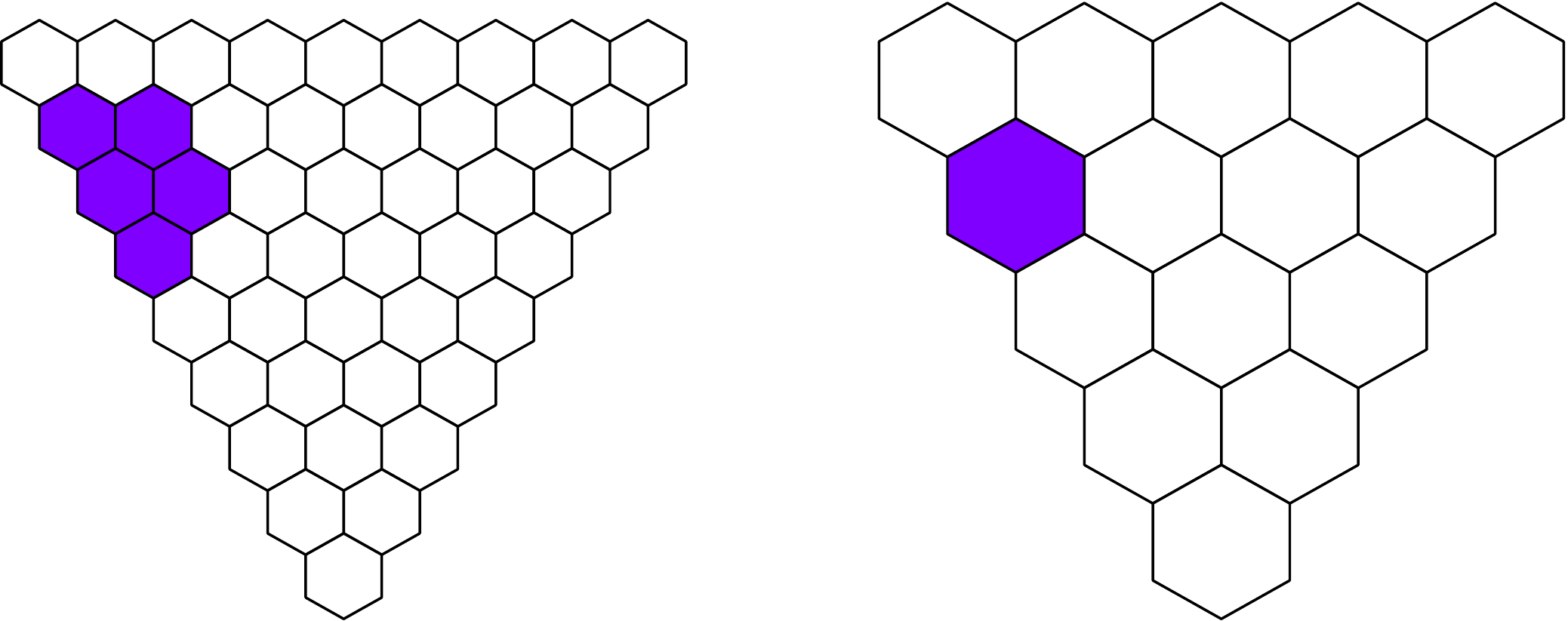}\\
    \vspace{0.75em}
    \includegraphics[width=0.52\linewidth]{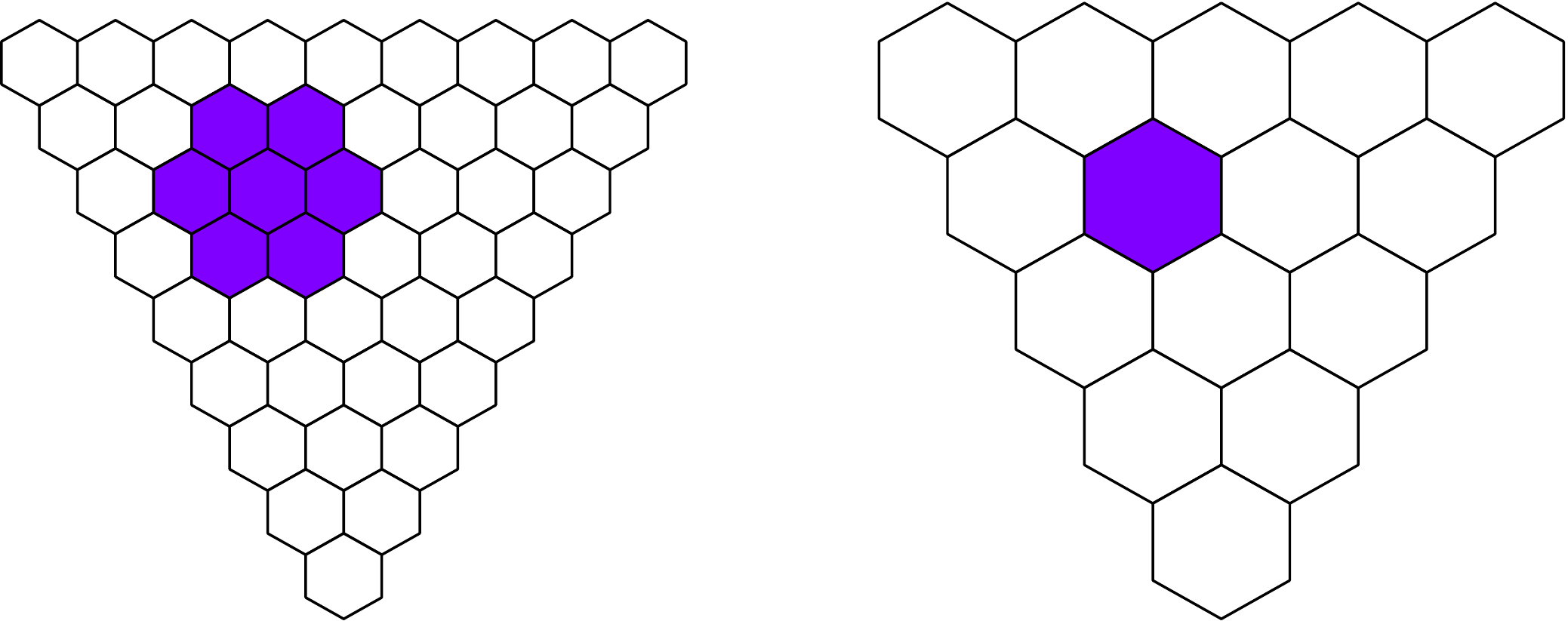}
    \caption{
        Two instances of a hexagonal pooling filter in action. The colored pixels on the left are being pooled over to produce the colored pixel on the right.
    }
    \label{fig:hexa_pool}
\end{figure}

\section{Details of Experiments}\label{sec:details}

We implemented our models with PyTorch. We use the Adam optimizer in all experiments. The width of the platonic pixelizations are chosen such that pooling can be applied three times and the number of pixels is comparable to previous works (e.g., \citet{jiang2019spherical}).

\subsection{Classification}\label{sec:details_classification}

We use the CNN architecture of \cref{sec:arch_class} for classification. Hyperparameters \texttt{\textcolor{purple}{C}} and \texttt{\textcolor{magenta}{F}} (defined in \cref{sec:arch}) are set to $20$ and $0.25$ respectively. Rate of dropout is set to $0.333$. We use a batch size of $128$. The model is trained for $50$ epochs with an initial learning rate of $10^{-3}$ that is multiplied by $0.1$ every $20$ epochs. Training a model takes near 3 hours on a V100 GPU.

\subsection{Segmentation}\label{sec:details_segmentation}

We use the U-Net architecture described in \cref{sec:arch_segm} for the two segmentation tasks.
The models were trained with an initial learning rate of $3 \cdot 10^{-3}$ that is multiplied by $0.7$ every $20$ epochs.

\subsubsection{Stanford2D3DS}

The hyperparameters \texttt{\textcolor{purple}{C}} and \texttt{\textcolor{magenta}{F}} were set to $16$ and $0.2$ respectively. We used a batch size of $16$ and a dropout rate of $0.333$. The model was trained for $120$ epochs. Training a model the three cross-validation splits takes near 4 hours on a Quadro RTX 8000 GPU.

\subsubsection{HAPPI20}

The class weights used in the weighted cross-entropy loss are $0.00766805$, $0.94184578$, and $0.05048618$ for BG, TC, and AR respectively. The hyperparameter \texttt{\textcolor{purple}{C}} was set to $28$ and \texttt{\textcolor{magenta}{F}} was set to $0.5$. We used a batch size of $64$ and a dropout rate of $0.1$. The model is trained until no improvement in validation performance is observed for $30$ epochs. Training a model takes near 3 days on two V100 GPUs.

\newpage
\onecolumn
\section{Neural Network Architecture}\label{sec:arch}

The layers inside \texttt{Sequential} are applied sequentially. \texttt{SphereLayer} is the hierarchical sphere layer introduced in this paper. It consists of \texttt{GroupConv} and \texttt{PoolPolyBroadcast} which pools each grid, applies an $\Hg$-equivariant layer, and broadcasts the result back on the grids. \texttt{PolyPad} is the equivariant padding layer described in \cref{sec:padding}.

\subsection{Classification}\label{sec:arch_class}
\begin{Verbatim}[commandchars=\\\{\},fontsize=\small]
Sequential(
  PolyPad(poly=POLY, padding=1)
  GroupConv(in_channels=1, out_channels=\textcolor{purple}{C}, kernel_size=3)
  AssertShape(BATCH_SIZE, NUM_FACES, \textcolor{purple}{C}, NUM_FLAGS_PER_FACE, WIDTH, WIDTH)
  BatchNorm, ReLU, Dropout
  PolyPad(poly=POLY, padding=1)
  GroupConv(in_channels=\textcolor{purple}{C}, out_channels=\textcolor{purple}{C}, kernel_size=3)
  BatchNorm, ReLU, Dropout
  PolyPad(poly=POLY, padding=1)
  SphereLayer(
    PoolPolyBroadcast(poly=POLY, in_channels=\textcolor{purple}{C} * \textcolor{magenta}{F}, out_channels=\textcolor{purple}{C} * \textcolor{magenta}{F})
    GroupConv(in_channels=\textcolor{purple}{C}, out_channels=\textcolor{purple}{C}, kernel_size=3)
  )
  MaxPool(kernel_size=2, stride=2)
  GroupConv(in_channels=\textcolor{purple}{C}, out_channels=2 * \textcolor{purple}{C}, kernel_size=1)
  BatchNorm, ReLU, Dropout
  PolyPad(poly=POLY, padding=1)
  GroupConv(in_channels=2 * \textcolor{purple}{C}, out_channels=2 * \textcolor{purple}{C}, kernel_size=3)
  BatchNorm, ReLU, Dropout
  PolyPad(poly=POLY, padding=1)
  SphereLayer(
    PoolPolyBroadcast(poly=POLY, in_channels=2 * \textcolor{purple}{C} * \textcolor{magenta}{F}, out_channels=2 * \textcolor{purple}{C} * \textcolor{magenta}{F})
    GroupConv(in_channels=2 * \textcolor{purple}{C}, out_channels=2 * \textcolor{purple}{C}, kernel_size=3)
  )
  MaxPool(kernel_size=2, stride=2)
  GroupConv(in_channels=2 * \textcolor{purple}{C}, out_channels=4 * \textcolor{purple}{C}, kernel_size=1)
  BatchNorm, ReLU, Dropout
  GroupConv(in_channels=4 * \textcolor{purple}{C}, out_channels=4 * \textcolor{purple}{C}, kernel_size=3)
  BatchNorm, ReLU, Dropout
  PolyPad(poly=POLY, padding=1)
  SphereLayer(
    PoolPolyBroadcast(poly=POLY, in_channels=4 * \textcolor{purple}{C} * \textcolor{magenta}{F}, out_channels=4 * \textcolor{purple}{C} * \textcolor{magenta}{F})
    GroupConv(in_channels=4 * \textcolor{purple}{C}, out_channels=4 * \textcolor{purple}{C}, kernel_size=3)
  )
  MaxPool(kernel_size=2, stride=2)
  GroupConv(in_channels=4 * \textcolor{purple}{C}, out_channels=8 * \textcolor{purple}{C}, kernel_size=1)
  BatchNorm, ReLU, Dropout
  PolyPad(poly=POLY, padding=1)
  GroupConv(in_channels=8 * \textcolor{purple}{C}, out_channels=8 * \textcolor{purple}{C}, kernel_size=3)
  BatchNorm, ReLU, Dropout
  PolyPad(poly=POLY, padding=1)
  SphereLayer(
    if \textcolor{magenta}{F} > 0: PoolPolyBroadcast(poly=POLY, in_channels=8 * \textcolor{purple}{C}, out_channels=NUM_CLASSES)
    GroupConv(in_channels=8 * \textcolor{purple}{C}, out_channels=NUM_CLASSES, kernel_size=3)
  )
  AssertShape(BATCH_SIZE, NUM_FACES, NUM_CLASSES, NUM_FLAGS_PER_FACE, WIDTH / 8, WIDTH / 8)
  GlobalPool(dim=[1, 3, 4, 5], pool_fn=mean)
)
\end{Verbatim}

\subsection{Segmentation}\label{sec:arch_segm}
\texttt{UBlock} is similar to \texttt{Sequential} except that its output is concatenated with its input.
\begin{Verbatim}[commandchars=\\\{\},fontsize=\small]
Sequential(
  UBlock(
    PolyPad(poly=POLY, padding=1)
    GroupConv(in_channels=NUM_INPUT_CHANNELS, out_channels=\textcolor{purple}{C}, kernel_size=3)
    AssertShape(BATCH_SIZE, NUM_FACES, \textcolor{purple}{C}, NUM_FLAGS_PER_FACE, WIDTH, WIDTH)
    BatchNorm, ReLU, Dropout
    PolyPad(poly=POLY, padding=1)
    GroupConv(in_channels=\textcolor{purple}{C}, out_channels=\textcolor{purple}{C}, kernel_size=3)
    BatchNorm, ReLU, Dropout
    PolyPad(poly=POLY, padding=1)
    SphereLayer(
      PoolPolyBroadcast(poly=POLY, in_channels=\textcolor{purple}{C} * \textcolor{magenta}{F}, out_channels=\textcolor{purple}{C} * \textcolor{magenta}{F})
      GroupConv(in_channels=\textcolor{purple}{C}, out_channels=\textcolor{purple}{C}, kernel_size=3)
    )
    UBlock(
      MaxPool(kernel_size=2, stride=2)
      GroupConv(in_channels=\textcolor{purple}{C}, out_channels=2 * \textcolor{purple}{C}, kernel_size=1)
      BatchNorm, ReLU, Dropout
      PolyPad(poly=POLY, padding=1)
      GroupConv(in_channels=2 * \textcolor{purple}{C}, out_channels=2 * \textcolor{purple}{C}, kernel_size=3)
      BatchNorm, ReLU, Dropout
      PolyPad(poly=POLY, padding=1)
      SphereLayer(
        PoolPolyBroadcast(poly=POLY, in_channels=2 * \textcolor{purple}{C} * \textcolor{magenta}{F}, out_channels=2 * \textcolor{purple}{C} * \textcolor{magenta}{F})
        GroupConv(in_channels=2 * \textcolor{purple}{C}, out_channels=2 * \textcolor{purple}{C}, kernel_size=3)
      )
      UBlock(
        MaxPool(kernel_size=2, stride=2)
        GroupConv(in_channels=2 * \textcolor{purple}{C}, out_channels=4 * \textcolor{purple}{C}, kernel_size=1)
        BatchNorm, ReLU, Dropout
        PolyPad(poly=POLY, padding=1)
        GroupConv(in_channels=4 * \textcolor{purple}{C}, out_channels=4 * \textcolor{purple}{C}, kernel_size=3)
        BatchNorm, ReLU, Dropout
        PolyPad(poly=POLY, padding=1)
        SphereLayer(
          PoolPolyBroadcast(poly=POLY, in_channels=4 * \textcolor{purple}{C} * \textcolor{magenta}{F}, out_channels=4 * \textcolor{purple}{C} * \textcolor{magenta}{F})
          GroupConv(in_channels=4 * \textcolor{purple}{C}, out_channels=4 * \textcolor{purple}{C}, kernel_size=3)
        )
        UBlock(
          MaxPool(kernel_size=2, stride=2)
          GroupConv(in_channels=4 * \textcolor{purple}{C}, out_channels=8 * \textcolor{purple}{C}, kernel_size=1)
          BatchNorm, ReLU, Dropout
          PolyPad(poly=POLY, padding=1)
          GroupConv(in_channels=8 * \textcolor{purple}{C}, out_channels=8 * \textcolor{purple}{C}, kernel_size=3)
          BatchNorm, ReLU, Dropout
          PolyPad(poly=POLY, padding=1)
          SphereLayer(
            PoolPolyBroadcast(poly=POLY, in_channels=8 * \textcolor{purple}{C} * \textcolor{magenta}{F}, out_channels=8 * \textcolor{purple}{C} * \textcolor{magenta}{F})
            GroupConv(in_channels=8 * \textcolor{purple}{C}, out_channels=8 * \textcolor{purple}{C}, kernel_size=3)
          )
          Upsample(scale_factor=2, mode=NEAREST)
          GroupConv(in_channels=8 * \textcolor{purple}{C}, out_channels=4 * \textcolor{purple}{C}, kernel_size=1)
        )
        BatchNorm, ReLU, Dropout
        PolyPad(poly=POLY, padding=1)
        GroupConv(in_channels=8 * \textcolor{purple}{C}, out_channels=4 * \textcolor{purple}{C}, kernel_size=3)
        BatchNorm, ReLU, Dropout
        PolyPad(poly=POLY, padding=1)
        SphereLayer(
          PoolPolyBroadcast(poly=POLY, in_channels=4 * \textcolor{purple}{C} * \textcolor{magenta}{F}, out_channels=4 * \textcolor{purple}{C} * \textcolor{magenta}{F})
          GroupConv(in_channels=4 * \textcolor{purple}{C}, out_channels=4 * \textcolor{purple}{C}, kernel_size=3)
        )
        Upsample(scale_factor=2, mode=NEAREST)
        GroupConv(in_channels=4 * \textcolor{purple}{C}, out_channels=2 * \textcolor{purple}{C}, kernel_size=1)
      )
      BatchNorm, ReLU, Dropout
      PolyPad(poly=POLY, padding=1)
      GroupConv(in_channels=4 * \textcolor{purple}{C}, out_channels=2 * \textcolor{purple}{C}, kernel_size=3)
      BatchNorm, ReLU, Dropout
      PolyPad(poly=POLY, padding=1)
      SphereLayer(
        PoolPolyBroadcast(poly=POLY, in_channels=2 * \textcolor{purple}{C} * \textcolor{magenta}{F}, out_channels=2 * \textcolor{purple}{C} * \textcolor{magenta}{F})
        GroupConv(in_channels=2 * \textcolor{purple}{C}, out_channels=2 * \textcolor{purple}{C}, kernel_size=3)
      )
      Upsample(scale_factor=2, mode=NEAREST)
      GroupConv(in_channels=2 * \textcolor{purple}{C}, out_channels=\textcolor{purple}{C}, kernel_size=1)
    )
    BatchNorm, ReLU, Dropout
    PolyPad(poly=POLY, padding=1)
    GroupConv(in_channels=2 * \textcolor{purple}{C}, out_channels=2 * \textcolor{purple}{C}, kernel_size=3)
    BatchNorm, ReLU, Dropout
    PolyPad(poly=POLY, padding=1)
    SphereLayer(
      PoolPolyBroadcast(poly=POLY, in_channels=2 * \textcolor{purple}{C} * \textcolor{magenta}{F}, out_channels=2 * \textcolor{purple}{C} * \textcolor{magenta}{F})
      GroupConv(in_channels=2 * \textcolor{purple}{C}, out_channels=2 * \textcolor{purple}{C}, kernel_size=3)
    )
    AssertShape(BATCH_SIZE, NUM_FACES, 2 * \textcolor{purple}{C}, FLAGS_PER_FACE, WIDTH, WIDTH)
    GlobalPool(dim=[3], pool_fn=mean)
  )
  BatchNorm, ReLU, Dropout
  Conv(in_channels=2 * \textcolor{purple}{C} + NUM_INPUT_CHANNELS, out_channels=NUM_CLASSES, kernel_size=1)
)
\end{Verbatim}

\section{Orbit-Finding Algorithm}

\begin{algorithm}
    \caption{Orbit-Finding for Parameter-Sharing}
    \label{alg:1}
\begin{algorithmic}
    \STATE {\bfseries Input:} $a \in \set{A}$, generators $\Gg^* \subseteq \Gg$ with $\langle\Gg^* \rangle = \Gg$
    \STATE {\bfseries Output:} Orbit $\set{O} = a^{\Gg}$
    \STATE $\mathcal{S} := \mathrm{stack}(a)$
    \STATE $\set{O} := \{a\}$
    \WHILE{$\mathcal{S}$ is not empty}
    \STATE $b := \mathcal{S}.\mathrm{pop}()$
    \FOR{$\gg \in \Gg^*$}
    \IF{$\gg \cdot b \notin \set{O}$}
    \STATE $\set{O} := \set{O} \cup \{\gg \cdot b\}$
    \STATE $\mathcal{S}.\mathrm{push}(\gg \cdot b)$
    \ENDIF
    \ENDFOR
    \ENDWHILE
\end{algorithmic}
\end{algorithm}}{}

\end{document}